\documentclass[10pt,twocolumn,letterpaper]{article}
\usepackage[pagenumbers]{cvpr}

\usepackage{cuted} %

\usepackage{currfile} %

\usepackage{caption} %

\usepackage{amsmath, amssymb, amsthm}
\usepackage{algorithm, algpseudocode}
\usepackage{booktabs, array}
\usepackage[table]{xcolor}
\usepackage{tabularx}
\usepackage{comment}
\usepackage{tikz}
\usepackage{fix-cm}
\usepackage{pifont}



\usepackage{listings}
\usepackage[scaled=0.85]{DejaVuSansMono}
\definecolor{codegray}{gray}{0.45}
\definecolor{ghcomment}{HTML}{6E7781}
\definecolor{ghkeyword}{HTML}{CF222E}
\definecolor{ghstring}{HTML}{0A3069}
\definecolor{ghconst}{HTML}{0550AE}
\definecolor{ghfunc}{HTML}{8250DF}
\lstdefinestyle{torch}{
  language=Python,
  basicstyle=\fontencoding{T1}\ttfamily\fontsize{8pt}{9.6pt}\selectfont,
  commentstyle=\color{ghcomment},
  keywordstyle=\color{ghkeyword},
  stringstyle=\color{ghstring},
  morekeywords=[2]{zip,min}, keywordstyle=[2]\color{ghfunc},
  morekeywords=[3]{None},    keywordstyle=[3]\color{ghconst},
  literate=*{0}{{{\color{ghconst}0}}}1 {1}{{{\color{ghconst}1}}}1 {2}{{{\color{ghconst}2}}}1,
  showstringspaces=false,
  columns=fullflexible,
  keepspaces=true,
  breaklines=false,
  aboveskip=2pt, belowskip=2pt,
}

\makeatletter
\@ifpackageloaded{lineno}{%
  \RemoveFromHook{build/column/before}[lineno]%
  \AddToHook{build/column/after}{%
    \setbox\@outputbox\vbox{%
      \boxmaxdepth\@maxdepth
      \protected@write\@auxout{}{%
        \string\def\string\@LN@column{\if@firstcolumn1\else2\fi}}%
      \box\@outputbox}}%
}{}
\makeatother

\definecolor{cvprblue}{rgb}{0.21,0.49,0.74}
\usepackage[breaklinks,colorlinks,allcolors=cvprblue]{hyperref}

\newcommand{\sarr}{\mbox{$\,\to\,$}}
\newcommand{\mpar}[1]{\medskip\noindent\textbf{#1}}
\newcommand{\fpar}[1]{\noindent\textbf{#1}}

\title{
\vspace{-2.0em}
Linear Probes Can Guide: Training-Free Control of Representation Generators
}

\author{
    Pengwei Wang$^{1}$ \quad
    Jos\'e Morano$^{1,2}$ \quad
    Botond Fazekas$^{1,2}$ \quad
    Hrvoje Bogunovi\'c$^{1,2}$ \\[0.5em]
    $^{1}$ AI Institute, Center for Medical Data Science, Medical University of Vienna, Austria \\
    $^{2}$ Christian Doppler Lab for AI in Retina, Medical University of Vienna, Austria%
    \vspace{-40pt}%
}

\begin{document}

\maketitle
\begin{strip}
\centering
{\large {Project Page: \href{https://penway.cn/lism/}{\nolinkurl{penway.cn/lism}}}\ \hbox{}\par}\vspace*{30pt}
\includegraphics[width=1\textwidth]{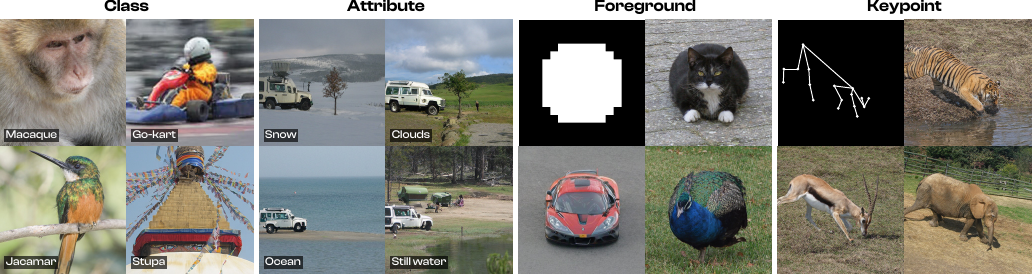}
\captionsetup{hypcap=false}
\captionof{figure}{\textbf{Linear Guidance in Semantic Mean-Space (LiSM)} achieves both global semantic control, such as class and attributes, and local control, such as foreground maps and keypoints, using only an ImageNet-trained generator, linear probes, and a fitted feature covariance.}
\captionsetup{hypcap=true}
\label{fig:teaser}
\end{strip}

\begin{abstract}
Representation-space generative models, such as Representation Autoencoders (RAEs), leverage pretrained vision foundation models (VFMs) primarily to accelerate training and improve sample fidelity.
Yet VFMs are more than reconstruction tokenizers, as they feature rich representation spaces that are highly organized and linearly readable.
We observe that this removes a key obstacle of post-hoc guidance: by linearity of expectation, a clean probe applied to an ideal denoiser's prediction yields the exact posterior mean of the probe scores at any noise level.
This regime, which we call Semantic Mean-Space, turns semantic guidance into a Bayesian linear inverse problem, bypassing the need for training-time conditioning.
With this foundation, we propose \textbf{LiSM}, a unified framework for \textbf{Linear Guidance in Semantic Mean-Space}.
To validate the framework, we developed and adapted different solvers, and characterized them based on how they estimate probe-space uncertainty and propagate the correction through the representation.
In addition, we propose Fit-Path, a solver designed to achieve strong global and local control. 
Our results in four control setups (class, attributes, foreground, and keypoints) show that LiSM methods, and Fit-Path in particular, provide effective, lightweight control across global and local tasks, outperforming noise-aware and pixel-space guidance.
\end{abstract}

\section{Introduction}
\label{sec:intro}

Current vision foundation models (VFMs) such as DINOv3~\cite{dinov3} and SigLIP~2~\cite{siglip2} present high generalizability to a wide range of tasks.
A defining property of these models is that they produce representations with strong linear readability~\cite{dinov2,dinov3}, so that single linear layers can be used to identify local and global attributes, and thus perform tasks such as classification and segmentation.

In recent years, VFMs are increasingly being incorporated into the generation process of generative models for accelerating training or improving generation quality.
Some relevant approaches include aligning intermediate states~\cite{repa,repae}, regularizing the variational autoencoder (VAE) tokenizer~\cite{vavae}, co-denoising~\cite{redi,reg}, or directly generating in representation space~\cite{rae,scalerae,raev2}.

Beyond fidelity, another challenge of generation is control, i.e., guiding the model to fulfill semantic requests~\cite{cfg,controlnet}.
The most common approach is training a conditional generator with paired data, but it limits conditions to those used for training.
To avoid this, post-hoc methods have been proposed.
Some of the most prominent train a noise-aware readout~\cite{cg,readoutguidance}, which requires retraining per condition.
Training-free alternatives apply off-the-shelf predictors to the decoded clean estimate~\cite{ug,freedom,tfg}, but decode and backpropagate through the predictor and denoiser at every step.
Linear inverse solvers~\cite{ddnm,pigdm,tmpd} need no readout network, but are mostly applied to pixel-level restoration.

In this work, we ask whether the linear readability of VFMs can be exploited for guidance in generative models operating in the representation space, as linear inverse solvers do at the pixel level.
This would avoid per-condition retraining and decoding, while preserving generalizability.
Our key observation is that, in this space, a linear readout trained on clean data remains valid at every noise level.
The reason is as follows.
An ideal flow denoiser inherently predicts the posterior mean in any space. In the representation space, the semantic readout is linear, and a linear operator commutes with the expectation; thus, applying a linear probe, trained on clean data, to the denoiser's prediction yields the exact posterior mean of the semantic scores (the probe outputs) at any noise level on the flow trajectory.
We call this regime \textbf{Semantic Mean-Space}.

This regime opens up a new opportunity of how generative models can be guided at inference time.
In pixel or VAE spaces, a nonlinear readout does not commute with the expectation, so, applied to the clean prediction, it does not give the expected score that guidance needs to weight a request.
In Semantic Mean-Space, the linear probe does commute, and global attribute steering and dense spatial control naturally become Bayesian linear inverse problems, for which solvers such as DDNM~\cite{ddnm}, $\Pi$GDM~\cite{pigdm}, and DTMPD~\cite{tmpd} provide lightweight, training-free conditioning.

However, linear semantic guidance in a representation space
differs in a crucial way from restoration in pixel space: a linear probe measures only a low-dimensional projection of the feature tokens, leaving unprobed channels unconstrained.
Pushing a sample exclusively along the probe normal can easily satisfy the readout in latent space without producing the corresponding visual change after decoding.
We identify this failure mode as \emph{probe hacking}.

To understand what makes linear semantic guidance succeed, we propose a unified framework: \textbf{Li}near guidance in \textbf{S}emantic \textbf{M}ean-space (\textbf{LiSM}). Our main contributions are as follows:
\begin{itemize}
  \item We show that representation flows admit exact posterior-mean semantic readouts from clean linear probes, which turns semantic guidance into a Bayesian linear inverse problem, giving rise to LiSM.
  \item We describe the LiSM framework along two axes, request weighting and representation response, and use it to adapt existing solvers and explore new solver design points.
  \item We propose Fit-Path, a LiSM solver that achieves consistently strong performance across global and local tasks.
  \item We evaluate different guidance frameworks on class, attribute, foreground, and keypoint control with an RAE generator, showing that the best LiSM solvers match or exceed pixel-space and noise-aware guidance in request adherence on all tasks.
\end{itemize}

\section{Related Work}

\fpar{Representation-space generative models.}
Diffusion models typically operate in pixel space~\cite{ddpm, iddpm} or in the latent space of a reconstruction-oriented VAE~\cite{ldm}, whose features are poorly suited to linear readout~\cite{rae}. Recent work brings pretrained representations into generation by aligning intermediate denoiser features with them~\cite{repa,repae}, regularizing the tokenizer~\cite{vavae}, jointly denoising images and representations~\cite{redi,reg}, or generating directly in the representation space of a frozen VFM~\cite{dinov2,dinov3,siglip,siglip2}, as in RAE~\cite{rae}, SVG~\cite{svg}, Scale-RAE~\cite{scalerae}, and RAEv2~\cite{raev2}.
In such representation-space generators, the \emph{generated variable itself} is linearly readable.
Representation readability has been used to select samples~\cite{scalerae} and to edit conditioning embeddings~\cite{didae}. Linear structure is also widely used to read and steer the \emph{internal activations} of large language models~\cite{iti,linearrep,sae} and diffusion models~\cite{asyrp,beyondsurface,sdxlsae}.
However, to our knowledge, linear probes fitted on VFM features have not been used to guide sampling.

\mpar{Guidance with learned readouts.}
Training-time conditioning, such as classifier-free guidance (CFG)~\cite{cfg} and spatial adapters~\cite{controlnet,t2iadapter,gligen}, requires paired data and retraining for every new control modality. Inference-time guidance instead keeps the generator fixed and steers it with an external readout. Classifier Guidance (CG)~\cite{cg} needs a classifier trained on inputs corrupted by the generator's own noise process, so it is tied to that diffusion's schedule and state space. Training-free methods reuse off-the-shelf predictors on the decoded clean estimate~\cite{freedom,ug,tfg}, but every guided step decodes to pixels and backpropagates through both the predictor and the denoiser, which Universal Guidance (UG) repeats several times per step~\cite{ug}. Such guidance is also unreliable on the unseen noisy estimates of early sampling~\cite{tfg} and prone to adversarial gradients~\cite{shen2024understanding}.

\mpar{Inverse-problem solvers and posterior approximations.}
Several solvers for inverse problems, with no need for an auxiliary network, were built for pixel-level restoration, differing mainly in how they approximate the clean posterior.
DPS~\cite{dps} and AdamDPS~\cite{adamdps} use a point mass at the clean estimate and backpropagate through the denoiser. DDiff~\cite{ddiff} carries a dual variable and DCPS~\cite{dcps} runs inner variational and Langevin loops. Projection methods~\cite{ddrm,ddnm,diffpir} enforce the linear constraint on the clean estimate. 
In contrast, Gaussian moment solvers add a posterior covariance: isotropic in $\Pi$GDM~\cite{pigdm}, DMPS~\cite{dmps}, and CoDPS~\cite{codps}, estimated from denoiser Jacobians in DTMPD~\cite{tmpd}, and initialized from data and updated online in Free Hunch~\cite{freehunch}.
All target a measurement likelihood, mostly linear-Gaussian, rather than semantic requests.
Closer to our setting, for the special case of an exponential tilt of a linear readout, the exact conditional mean is obtained by evaluating the denoiser once at a shifted input~\cite{moitra}.
In \emph{Semantic Mean-Space}, a clean linear probe acts as a linear operator on the denoiser output, without pixel decoding or readout backpropagation, so these solvers become applicable to semantic control. 
Our \textbf{LiSM} framework adapts and unifies these solvers for representation flows along two axes: request weighting and representation response.

\section{Preliminary}\label{sec:prelim}

\fpar{Representation flow models.}
A representation autoencoder~\cite{rae} pairs a frozen, pretrained encoder $E:\mathcal{X}\to\mathbb{R}^{d}$ with a decoder $D:\mathbb{R}^{d}\to\mathcal{X}$ trained to invert it, where $\mathcal{X}$ is the image space.
The encoder produces representation $z=E(x)\in \mathbb{R}^{d}$ and the decoder, the reconstructed image $\hat{x}=D(z)$.
The flow path, with $t=1$ pure noise and $t=0$ clean data, is formulated as:
\begin{equation}
    z_t = (1-t)z+t\varepsilon, \quad \varepsilon \sim \mathcal{N}(0,I).
\end{equation}
The denoiser $f_\theta(z_t,t)$ predicts the clean representation $\hat{z}$. An ideal, MSE-optimal denoiser predicts the posterior mean~\cite{ManorMichaeli}, where $Z$ is the random clean representation:
\begin{equation}
    \hat{z}=f_\theta(z_t,t)=\mathbb{E}[Z\mid z_t].
\end{equation}
Sampling an image without guidance integrates the velocity given by this mean prediction,
\begin{equation}
    v_\theta(z_t,t)=\frac{z_t-\hat{z}}{t}
\end{equation}
with an ODE backward from $t=1$ to $t=0$, starting from $z_1\sim\mathcal{N}(0,I)$ and decoding $z_0$.

\mpar{Guidance as conditional sampling.}
Let $h:\mathbb{R}^{d}\to\mathbb{R}^{q}$ be a readout of a property of the clean representation, e.g., a classifier, a segmentor, or an image-space network $h_{\mathrm{img}}$ through $h(z)=h_{\mathrm{img}}(D(z))$.
Moreover, let $L:\mathbb{R}^{q}\to[0,\infty)$ be a weight that states how acceptable a readout value is.
Guidance aims to sample from the target
\begin{equation}
    p_L(z)\propto p(z)\,L\big(h(z)\big),
\end{equation}
where $p(z)$ is the unconditional distribution of representations. This covers likelihood conditioning~\cite{tds} and reward tilting~\cite{fksteering}. At a noisy state $z_t$ the clean representation is uncertain, so the relevant quantity is the expected weight
\begin{equation}
    \pi_t(z_t)=\mathbb{E}\big[L(h(Z))\mid z_t\big].
\end{equation}
Classifier guidance~\cite{cg} learns it with a noise-conditioned classifier, whereas DPS~\cite{dps} and UG~\cite{ug} approximate it by evaluating $L(h(\cdot))$ at the clean prediction $\hat z$.

\mpar{Guidance as a correction of the clean prediction.}
Sampling from the target means following its own flow, which uses the clean prediction under the target, $\hat z_L=\mathbb{E}_{p_L}[Z\mid z_t]$, in place of $\hat z$. Since the target's noisy marginal is proportional to $p_t(z_t)\,\pi_t(z_t)$, where $p_t$ is the density of $z_t$, Tweedie's formula~\cite{efron2011} shows that for an ideal denoiser the two differ by
\begin{equation}
    \Delta:=\hat z_L-\hat z=c_t\,\nabla_{z_t}\log\pi_t(z_t),\qquad c_t=\frac{t^2}{1-t},
\end{equation}
for $t\in(0,1)$. Guidance is therefore a correction of the clean prediction, which enters the velocity as $v_L=(z_t-\hat z-\Delta)/t$. Solvers differ in how they estimate $\Delta$ from the readout and the denoiser.

\section{Linear Guidance in Semantic Mean-Space}
\label{sec:lism_method}

Guidance needs a correction of the clean prediction. Most methods obtain it by decoding the prediction and reading it with a pixel-space network, or by training a noise-aware readout. However, in a representation flow neither is needed: the generated variable is the representation itself, and a frozen encoder organizes it so that attributes are linearly readable. At the same time, at an intermediate step the clean prediction is not a clean sample but a posterior average of possible representations. Let $r(z)=Wz+b$, with $W\in\mathbb{R}^{m\times d}$ and $b\in\mathbb{R}^{m}$, be a linear semantic probe fitted on clean representations and kept frozen. Our key observation is that it commutes with the posterior mean, so it reads the (ideal) denoiser's output exactly, giving the posterior mean of the clean semantic scores at every noise level:
\begin{equation}
    r(\mathbb{E}[Z\mid z_t])=\mathbb{E}[r(Z)\mid z_t].
\end{equation}
This follows from the linearity of conditional expectation. We call this regime \emph{Semantic Mean-Space}; it reduces semantic guidance to a problem on a few linear scores with known means, which linear inverse-problem solvers can handle. The identity requires the readout to be linear in $z$ (with bias), so we apply probes at the logit level.

\begin{figure}[!t]
    \includegraphics[width=1\linewidth]{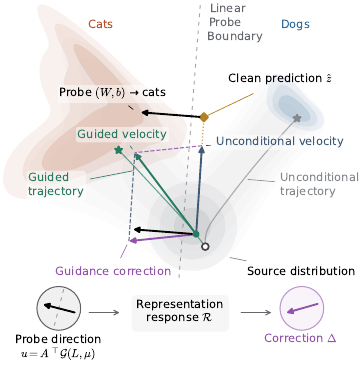}
    \caption{\textbf{Illustration of LiSM.} VFMs organize representations into a linearly readable space. 
    The unconditional flow denoiser predicts the \emph{Clean prediction}, and a linear probe reads semantic attributes from this intermediate prediction (\emph{Probe $\rightarrow$ cats}).
    A solver can then use this probe direction with \emph{representation response} to compute the correction and steer the \emph{guided trajectory}.}
    \label{fig:lism}
\end{figure}

Fig.~\ref{fig:lism} and Alg.~\ref{alg:lism} summarize the LiSM framework as a four-step procedure that modifies the original flow velocity through semantic guidance.
First, given the clean prediction $\hat z$, the linear probe reads the mean scores $\mu$ using a request operator $(A,c)$ built from the probe.
Second, a weighting rule $\mathcal{G}$ determines how these scores should be weighted and produces request weights $\lambda$, which are used together with $A$ to get the probe direction $u$.
Third, a response rule $\mathcal{R}$ translates this direction into a representation correction $\Delta=\alpha \mathcal{R}(u)$, where $\alpha\ge0$ controls the guidance strength.
Finally, the guided velocity $v_L$ is used for the update.

We next describe how global and local semantic requests can be expressed through a common linear interface, followed by the design choices underlying different solvers.

\subsection{Semantic Requests}\label{sec:requests}

A semantic request states a condition on linear probe scores. It consists of a specification $s$ (e.g., a target class, attribute signs, a foreground mask, or keypoints) and the weight function $L$, and defines the target $p_L(z)\propto p(z)\,L\big(y(z)\big)$ of Sec.~\ref{sec:prelim} (readout $h=y$). The readout is the request operator, which maps a representation to $q$ request scores
\begin{equation}
    y(z)=Az+c,\quad y_i(z)=a_i^\top z+c_i,\quad i=1,\ldots,q,
\end{equation}
each defined so that $y_i(z)\ge0$ means requirement $i$ is satisfied. The operator $(A,c)$ is built from the trained probe $(W,b)$ and $s$ by selecting rows, taking differences (class margins), applying signs and thresholds, and pooling, and is fixed before sampling. A request may impose hard constraints or soft preferences, on global attributes or on local spatial properties.

\begin{algorithm}[!t]
\caption{LiSM Framework}
\label{alg:lism}
\small
\textbf{Inputs:} Denoiser $f_\theta$, decoder $D$, request operator $(A,c)$ built from the probe $(W,b)$,
weight function $L$, weighting rule $\mathcal{G}$, response rule $\mathcal{R}$ and guidance strength $\alpha$.

\medskip
\noindent At each velocity evaluation of the solver (given $z_t$ and $t$):\\[2pt]
$\begin{alignedat}{2}
\hat{z} &= f_\theta(z_t,t) &\quad& \triangleright\ \text{\textbf{(1)} Clean prediction}\\
\mu &= A\hat{z}+c && \triangleright\ \text{\textbf{(1)} Mean scores}\\
\lambda &= \mathcal{G}(L,\mu) && \triangleright\ \text{\textbf{(2)} Request weights}\\
u &= A^\top \lambda && \triangleright\ \text{\textbf{(2)} Probe direction}\\
\Delta &= \alpha \mathcal{R}(u) && \triangleright\ \text{\textbf{(3)} Representation correction}\\
v_L &= \frac{z_t-(\hat{z}+\Delta)}{t} && \triangleright\ \text{\textbf{(4)} Guided velocity}\\[3pt]
\noalign{\noindent\text{Sampling:}}
z_1 &\sim \mathcal{N}(0,I), && \\
z_0 &= \rlap{$\operatorname{ODESolve}\!\left(v_L,z_1,[1,0]\right),$} && \\
\hat{x} &= D(z_0). &&
\end{alignedat}$\\[4pt]
\footnotesize The rules $\mathcal{G}$ and $\mathcal{R}$ may use time $t$, uncertainty estimates, and the current denoiser state $(z_t, \hat{z})$; these dependencies are suppressed for readability. A direct correction uses the identity response, $\mathcal{R}(u)=u$.
\end{algorithm}

Global requests apply these rows to spatially pooled features, for example to a positive class margin against negative classes. Local requests apply them to individual tokens, for example to require foreground scores above a threshold inside a mask and below it outside. We combine the requirements through
\begin{equation}
    L(y) = \prod^q_{i=1}\ell_i(y_i), \quad\ell_i(y)=
    \begin{cases}
    1, & y\ge0\\
    \beta_i, & y<0
    \end{cases}\label{eq:likelihood}
    ~,
\end{equation}
where $\beta_i=0$ imposes a hard constraint and $0<\beta_i<1$ assigns a soft penalty to its violation.

\subsection{Weighting Rules}\label{sec:weighting}
The clean prediction $\hat z$ gives the mean scores $\mu=A\hat z+c$. A weighting rule $\mathcal{G}$ produces the request weights
\begin{equation}
    \lambda = \mathcal{G}(L,\mu).
\end{equation}
These specify how the request operator contributes to the probe direction $u=A^\top \lambda$. The construction of $\lambda$ depends on the solver and may incorporate posterior uncertainty. 

The posterior uncertainty is useful as the means alone do not determine how likely the request is to be satisfied. Since the weight function $L$ is generally nonlinear, conditioning requires its expectation over the possible clean representations (Sec.~\ref{sec:prelim}):
\begin{equation}
\pi_t(z_t)=\mathbb{E}[L(Y)\mid z_t], \quad Y=y(Z)=AZ+c.
\end{equation}
Different solvers approximate this quantity differently.

\mpar{Point weighting.}
Point-evaluation methods~\cite{dps,adamdps,ddiff} apply a loss or request weight directly to the mean prediction,
\begin{equation}
    \mathcal G_{\mathrm{Point}}(L,\mu)=\nabla_\mu\log\tilde L(\mu),
\end{equation}
where $\tilde L$ is a differentiable surrogate of $L$.
This ignores posterior uncertainty and does not in general recover $\pi_t$.

\mpar{Gaussian weighting.}
Uncertainty-aware methods also model the variation of the possible clean scores.
In Gaussian approximations~\cite{dcps,codps,pigdm,tmpd}, the clean scores are modeled by a Gaussian $Y\mid z_t\sim\mathcal{N}(\mu,S_t)$, where covariance follows from the same linearity as mean commutation:
\begin{equation}
    S_t=\mathrm{Cov}[Y\mid z_t]=A\,\Sigma_t\,A^\top,\quad \Sigma_t=\mathrm{Cov}[Z\mid z_t].
\end{equation}

The approximate request weight and its sensitivity to the mean scores are
\begin{equation}
\hat\pi_L(\mu,S_t)
=\mathbb E[L(Y)],
\quad
\mathcal{G}_\mathrm{Gauss}=\nabla_\mu\log\hat\pi_L(\mu,S_t),
\end{equation}
where $S_t$ is held fixed when differentiating.
The Gaussian variants approximate this covariance differently:
\begin{equation}
\begin{aligned}
    \mathcal G_{\mathrm{Iso}}(L,\mu)
    &=\mathcal G_{\mathrm{Gauss}}
      (L,\mu;\sigma_t^2AA^\top),\\
    \mathcal G_{\mathrm{Fit}}(L,\mu)
    &=\mathcal G_{\mathrm{Gauss}}
      (L,\mu;A\hat\Sigma_tA^\top).
\end{aligned}
\end{equation}
The isotropic variant~\cite{codps,pigdm} uses a scalar posterior-variance estimate
$\sigma_t^2$.
Instead, the fitted variant used in our method estimates a clean prior covariance
$\hat\Sigma_0$ and propagates it under a Gaussian-prior approximation:
\begin{equation}
    \hat\Sigma_t
    =\Big(\hat\Sigma_0^{-1}
      +\tfrac{(1-t)^2}{t^2}I\Big)^{-1}.\label{eq:cov-transfer}
\end{equation}
Other solvers use Jacobian-based uncertainty estimates, as in DTMPD~\cite{tmpd}, or inner variational inference, as in DCPS~\cite{dcps}; their solver-specific constructions are detailed in Appendix~\ref{append:solver-detail}.

\subsection{Representation Response}
With the weighted probe direction $u$, the final step is to turn it into a correction with representation response $\Delta=\alpha\mathcal{R}(u)$, which enters the guided velocity in Alg.~\ref{alg:lism}. Next, we introduce some response designs.
Some solvers with optimizer state or inner inference steps require additional update mechanisms beyond these response maps; their adaptations are detailed in Appendix~\ref{append:solver-detail}.

\mpar{Direct and Prior-Based Responses.}
The simplest choice is to directly apply the probe direction with identity, $\mathcal{R}_\mathrm{Id}(u) = u$, without considering dependencies between probed and unprobed directions.
Under a Gaussian posterior approximation with consistent Gaussian weighting, these dependencies enter through the covariance response:
\begin{equation}
    \mathcal{R}_\mathrm{Cov}(u) = \hat\Sigma_t u.
\end{equation}
Whereas $S_t$ informs request weighting, $\Sigma_t$ propagates the correction through representation dependencies and can be approximated analytically without additional denoiser evaluations.

\mpar{Denoiser-Based Responses.}
A fixed prior approximation may miss dependencies specific to the current state. For an ideal posterior-mean denoiser, the clean posterior covariance is available through its Jacobian:
\begin{equation}
    \Sigma_t = c_t J_{f_\theta},\quad c_t=\frac{t^2}{1-t}, \quad J_{f_\theta} = \frac{\partial f_\theta(z_t,t)}{\partial z_t}.
\end{equation}
The $J$ can be obtained through differentiation of the denoiser. In practice, reverse-mode differentiation of the network computes vector--Jacobian
product (VJP) $J_{f_\theta}^\top u$:
\begin{equation}
    \mathcal{R}_{\mathrm{VJP}}(u)=c_t J_{f_\theta}^\top u,
\end{equation}
which is widely used~\cite{dps,adamdps,pigdm,tmpd}. Alternatively, forward-mode differentiation computes Jacobian--vector product (JVP) $J_{f_\theta} u$:
\begin{equation}
    \mathcal{R}_{\mathrm{JVP}}(u)=c_t J_{f_\theta} u,
\end{equation}
which is less explored and used. 
For an ideal denoiser, these two should be the same (since $J$ is symmetric).

\mpar{\textit{Proposed}: Path-Integral Responses.}
The local response describes the denoiser's infinitesimal change along the probe direction $u$. However, this response may vary along the direction even within a very short horizon. If the response changes over the displacement, extrapolating its initial estimate can mislead the correction.
We therefore propose to replace the pointwise response with a normalized path integral:
\begin{equation}
\begin{split}
    \mathcal{R}_{\mathrm{Path}}(u)
    &= \frac{c_t}{\tau}\int_0^\tau J_{f_\theta}(z_t+s u,t)\,u\,ds \\
    &= \frac{c_t}{\tau}\big[f_\theta(z_t+\tau u,t)-\hat z\big].
\end{split}
\end{equation}
It needs one extra denoiser evaluation and no differentiation.
This considers the averaged response in a horizon $\tau$ along the probe direction. When the directional response is constant along the path, it equals the local response; otherwise, it incorporates changes in both response magnitude and direction.

Crucially, the local response is also recovered as $\tau\to0$, which finite-difference ($\mathcal{R}_\text{FD}$) approximations of the JVP use~\cite{graikos2026fast}. %
In our case, $\tau$ is purposely intended to be large to capture the average directional response along the query path.
As we will show, this improves global task request adherence, while keeping local adherence close to JVP.

Specifically, we control the path length through a maximum per-patch query radius $H$:
\begin{equation}
    \tau=\min\!\left\{c_t,\frac{H}{\max_p\|u_p\|_2}\right\},
\end{equation}
where $p$ indexes spatial patches, and a shared $\tau$ bounds every
patch's query displacement by $H$; the response is zero when $u=0$. %

\section{Experiments}

We compared our novel LiSM framework to two other image-guidance frameworks: pixel-space~\cite{ug,dps,adamdps,ddiff,dcps} and noise-aware~\cite{cg,adamdps}.
Then, we explored the solver design space, identifying and implementing additional solvers, and compared configurations with different weightings and responses, including Path-Integral Responses (Path).
Complementarily, we ported existing solvers~\cite{dcps,dps,adamdps,tmpd,pigdm,ddiff,freehunch,codps} to LiSM, and compared them against Fit-Path.
Finally, we evaluated the performance of Fit-Path using other VFMs beyond DINOv3~\cite{dinov3}: MAE~\cite{mae} and SigLIP~2~\cite{siglip2}.
Additional analyses of Path are presented in Appendix~\ref{append:exp}.

Unless stated otherwise, the experiments were performed on ImageNet~\cite{imagenet} with RAEv2~\cite{raev2} generator.
Specifically, we selected two global-level and two local-level guidance tasks to have a full view of the performance of different methods: class, image-level attributes, foreground and keypoints. For class, the denoiser receives the null class; for the other tasks, it is conditioned on the requested class with CFG scale 1. All methods share the same initial noise per request. We use 3{,}840 test requests in total; ablations use a fixed subset of 810.

Each task has an adherence and a preservation metric, both computed on decoded images by
external networks that differ from all guiding networks. Adherence is top-1 accuracy for class,
the fraction of images with a CLIP-based score above 0.1 for the requested attribute, IoU for
foreground, and PCK@0.2 for keypoints. Preservation is precision~\cite{precision} for class and, for the other tasks, the rate at which an external classifier still recognizes the requested class.
We also report hacking (Hack), which is the mean over tasks of internal minus external adherence, clipped at zero per task, where internal adherence is measured by the guiding predictor on the final sample.

For each method and task, we tune one hyperparameter, the guidance strength $\alpha$, on a development set of 384 requests disjoint from the test set; all other settings are fixed. We select by adherence for class, and by adherence subject to at least 75\% class preservation for the other tasks. This threshold is applied on the development set only, so preservation on the test set can be lower.
Details are in Appendix~\ref{append:imple-detail}.
Code will be made publicly available.

\subsection{LiSM vs. other guidance frameworks} 

\begin{figure*}[t]
    \centering
    \includegraphics[width=1\linewidth]{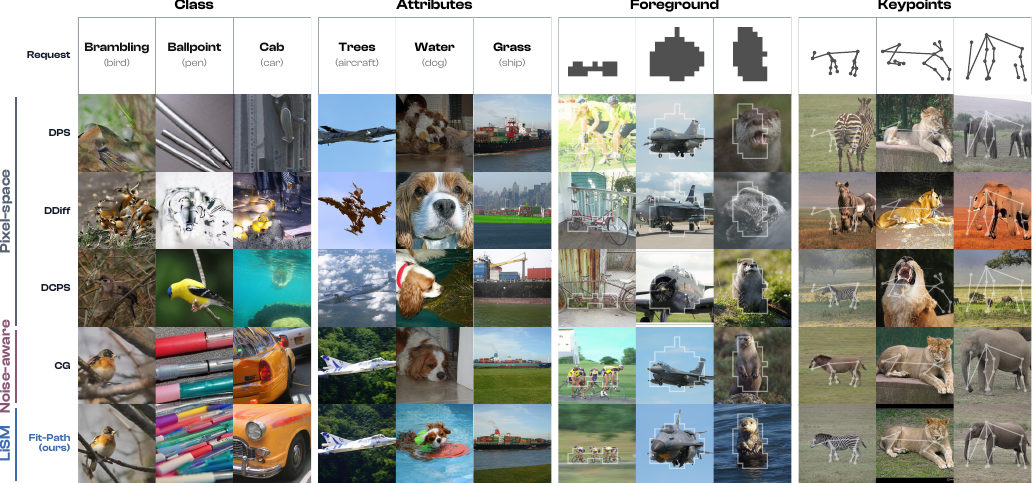}
    \caption{\textbf{Qualitative results for representative methods from three frameworks.} Images of the same column are starting from the same initial random state.}
    \label{fig:qualitative}
\end{figure*}

Table~\ref{tab:rae-main} shows the performance of the methods from different frameworks, including our LiSM solvers Fit-Id and Fit-Path, on the different evaluation tasks.
Both LiSM solvers use fitted Gaussian as weighting, but differ in the response: Fit-Id uses identity, while Fit-Path uses our proposed Path.
Qualitative examples of images generated by some of these methods are presented in Fig.~\ref{fig:qualitative}.
Full results for the different methods, with Adherence, Preservation, and Hacking are presented in Appendix~\ref{append:quant}, while additional qualitative examples are shown in Appendix~\ref{append:quali}.

\begin{table}[t]
\centering
\small
\begingroup
\definecolor{ctrlgray}{HTML}{808080}
\definecolor{RAEpixel}{HTML}{566274}
\definecolor{RAEnoise}{HTML}{8A5675}
\definecolor{RAElism}{HTML}{3E6CB0}
\setlength{\tabcolsep}{1.5pt}
\newcolumntype{z}[1]{>{\raggedleft\arraybackslash}p{#1pt}}
\newcommand{\gm}[2]{\mbox{#1\,{\scriptsize(#2)}}}
\begin{tabular}{l|z{36}|z{20}z{20}z{20}z{20}|z{20}|z{20}}
\textbf{Method} & \textbf{GM} & \textbf{Class} & \textbf{Attr.} & \textbf{Fg.} & \textbf{Kp.} & \textbf{Hack} & \textbf{Time} \\
 & {Adh.{(\scriptsize Pres.)}} & Acc. & CLIP & IoU & PCK & $\downarrow$ & s\,$\downarrow$ \\
\noalign{\hrule height 0.8pt}
\textcolor{ctrlgray}{\textit{Ref. (CFG)}} & \textcolor{ctrlgray}{--} & \textcolor{ctrlgray}{\textit{84.0}} & \textcolor{ctrlgray}{--} & \textcolor{ctrlgray}{--} & \textcolor{ctrlgray}{--} & \textcolor{ctrlgray}{--} & \textcolor{ctrlgray}{--} \\
\hline
\textcolor{ctrlgray}{\textit{Control}} & \textcolor{ctrlgray}{\gm{5.1}{81.4}} & \textcolor{ctrlgray}{0.1} & \textcolor{ctrlgray}{15.2} & \textcolor{ctrlgray}{27.4} & \textcolor{ctrlgray}{16.0} & \textcolor{ctrlgray}{--} & \textcolor{ctrlgray}{4.6} \\
\hline
\rowcolor{RAEpixel!12}
\multicolumn{8}{l}{\textcolor{RAEpixel}{\textbf{\textit{Pixel-space}}}} \\
UG\,\cite{ug} & \gm{32.4}{53.8} & 18.5 & 27.1 & 47.0 & 46.5 & 26.9 & 70.6 \\
DPS\,\cite{dps} & \gm{28.0}{61.7} & 21.2 & 19.8 & 38.4 & 38.4 & 18.1 & 22.5 \\
AdamDPS\,\cite{adamdps} & \gm{21.4}{51.1} & 10.7 & 19.6 & 31.6 & 31.6 & 28.1 & 22.5 \\
DDiff\,\cite{ddiff} & \gm{38.8}{55.5} & 31.0 & 35.8 & 38.6 & 52.7 & 31.6 & 16.5 \\
DCPS\,\cite{dcps} & \gm{18.9}{76.4} & 1.8 & 34.3 & \underline{66.4} & 31.4 & 16.8 & 76.6 \\
\hline
\rowcolor{RAEnoise!12}
\multicolumn{8}{l}{\textcolor{RAEnoise}{\textbf{\textit{Noise-aware}}}} \\
CG\,\cite{cg} & \gm{72.1}{78.0} & \underline{92.2} & 72.9 & 44.9 & \underline{89.8} & 7.2 & 5.0 \\
AdamCG\,\cite{adamdps} & \gm{65.0}{74.0} & 87.9 & 63.3 & 38.0 & 84.2 & 9.5 & 5.0 \\
\hline
\rowcolor{RAElism!12}
\multicolumn{8}{l}{\textcolor{RAElism}{\textbf{\textit{LiSM (ours)}}}} \\
Fit-Id & \gm{\underline{79.5}}{76.0} & \textbf{93.6} & \textbf{94.5} & 58.8 & 76.9 & 19.1 & 5.1 \\
\textbf{Fit-Path} & \gm{\textbf{87.2}}{72.8} & 90.2 & \underline{92.7} & \textbf{76.8} & \textbf{90.0} & 4.7 & 8.5 \\
\end{tabular}
\caption{\textbf{Guidance performance on four control tasks: class, attributes, foreground, and keypoints}.
GM: geometric mean of adherence and preservation over tasks.
\textbf{Bold}: best; \underline{underline}: second-best.
Hack: mean over tasks of internal minus external adherence, clipped at zero.
Control: generator without guidance toward the scored task (null for class; class-conditioned otherwise).
}
\label{tab:rae-main}
\endgroup
\end{table}

The performance of LiSM methods is overall superior to that of the \textit{Noise-aware} and \textit{Pixel-space} frameworks.
Across tasks, the best method in terms of geometric mean (GM) adherence is Fit-Path (87.2), followed by Fit-Id (79.5) and CG (72.1).
Fit-Path is also the method presenting the lowest hacking.
Interestingly, while Fit-Path provides the most balanced performance, Fit-Id performs particularly well on global tasks (Class and Attr.).
In general, the comparison to noise-aware methods shows that noise-aware training is not essential to representation space guidance, and linear probes with Semantic Mean-Space can produce very similar (e.g., the aircraft with trees in Fig.~\ref{fig:qualitative}) or better images (e.g., the dog in the water).

The improvements over pixel-space methods are even more pronounced, as they generally present much lower adherence than LiSM.
In addition, we observed that it is common, in methods such as DDiff, that the request is met but the image is not realistic (see Class, Fig.~\ref{fig:qualitative}).
On the other hand, DCPS shows overall good image quality, but low adherence.
In contrast, LiSM generates highly natural and reasonable images, while still following the request.

\subsection{Analyzing the LiSM design space}

To better understand what makes linear semantic guidance succeed, we explored the design space of linear inverse-problem solvers.
The results of the different LiSM-adapted solvers on the four evaluated tasks are shown in Tab.~\ref{tab:lism-ablation}.
For clarity, we add references, in the table, to solvers already studied in previous works for pixel-space inverse problems.

\begin{table}[t]
\centering
\small
\begingroup
\setlength{\tabcolsep}{1.25pt}
\newcolumntype{z}[1]{>{\raggedleft\arraybackslash}p{#1pt}}
\newlength{\lismresponsewidth}
\newlength{\lismresponsecandidate}
\settowidth{\lismresponsewidth}{Path}
\settowidth{\lismresponsecandidate}{Id~\cite{ddnm}}\ifdim\lismresponsecandidate>\lismresponsewidth\setlength{\lismresponsewidth}{\lismresponsecandidate}\fi
\settowidth{\lismresponsecandidate}{VJP~\cite{dps}}\ifdim\lismresponsecandidate>\lismresponsewidth\setlength{\lismresponsewidth}{\lismresponsecandidate}\fi
\settowidth{\lismresponsecandidate}{Cov~\cite{codps}}\ifdim\lismresponsecandidate>\lismresponsewidth\setlength{\lismresponsewidth}{\lismresponsecandidate}\fi
\settowidth{\lismresponsecandidate}{VJP~\cite{pigdm}}\ifdim\lismresponsecandidate>\lismresponsewidth\setlength{\lismresponsewidth}{\lismresponsecandidate}\fi
\newcommand{\response}[2]{\makebox[\lismresponsewidth][l]{#1\hfill#2}}
\begin{tabular*}{\linewidth}{@{}z{4}@{\extracolsep{\fill}}l@{\extracolsep{0pt}\hspace{2.5pt}}l|z{45}|z{45}|z{45}|z{24}@{\extracolsep{\fill}}z{4}@{}}
& $\mathcal G$ & $\mathcal R$ & \textbf{GM} & \textbf{Global} & \textbf{Local} & \textbf{Hack} & \\
 & & & Adh. {\scriptsize (Pres.) } & Adh. {\scriptsize (Pres.) } & Adh. {\scriptsize (Pres.) } & $\downarrow$ & \\
\noalign{\hrule height 0.8pt}
 & Point & \response{VJP}{\cite{dps}} & \mbox{81.0\,{\scriptsize (74.5)}} & \mbox{75.1\,{\scriptsize (63.9)}} & \mbox{\textbf{87.4}\,{\scriptsize (86.9)}} & 13.7 &  \\
 & Point & \response{Path}{} & \mbox{\underline{82.7}\,{\scriptsize (76.2)}} & \mbox{81.2\,{\scriptsize (72.3)}} & \mbox{84.2\,{\scriptsize (80.2)}} & 2.8 &  \\
\hline
 & Iso & \response{Cov}{\cite{codps}} & \mbox{50.1\,{\scriptsize (77.5)}} & \mbox{\textbf{94.0}\,{\scriptsize (68.1)}} & \mbox{26.7\,{\scriptsize (88.1)}} & 39.1 &  \\
 & Iso & \response{VJP}{\cite{pigdm}} & \mbox{59.7\,{\scriptsize (73.6)}} & \mbox{50.0\,{\scriptsize (64.9)}} & \mbox{71.4\,{\scriptsize (83.4)}} & 11.0 &  \\
 & Iso & \response{Path}{} & \mbox{74.1\,{\scriptsize (75.4)}} & \mbox{86.8\,{\scriptsize (65.9)}} & \mbox{63.2\,{\scriptsize (86.2)}} & 3.1 &  \\
\hline
 & Fit & \response{Id}{} & \mbox{79.0\,{\scriptsize (74.6)}} & \mbox{\underline{93.1}\,{\scriptsize (65.4)}} & \mbox{67.0\,{\scriptsize (85.1)}} & 19.7 &  \\
 & Fit & \response{JVP}{} & \mbox{71.3\,{\scriptsize (69.8)}} & \mbox{61.3\,{\scriptsize (58.9)}} & \mbox{83.0\,{\scriptsize (82.7)}} & 20.4 &  \\
 & Fit & \response{FD}{} & \mbox{73.2\,{\scriptsize (67.6)}} & \mbox{64.5\,{\scriptsize (55.0)}} & \mbox{83.0\,{\scriptsize (83.0)}} & 18.9 &  \\
 & Fit & \response{VJP}{} & \mbox{70.1\,{\scriptsize (69.3)}} & \mbox{58.1\,{\scriptsize (58.6)}} & \mbox{\underline{84.6}\,{\scriptsize (81.9)}} & 21.3 &  \\
 & Fit & \response{Path}{} & \mbox{\textbf{86.7}\,{\scriptsize (70.9)}} & \mbox{90.4\,{\scriptsize (61.1)}} & \mbox{83.2\,{\scriptsize (82.2)}} & 4.8 &  \\
\end{tabular*}
\caption{\textbf{Ablation of weighting $\mathcal G$ and response $\mathcal R$.}
Results in the ablation subset.
Geometric means (GM): over the four tasks, global over class and attributes, local over foreground and keypoints.
Iso, Fit: isotropic and fitted Gaussian.
Id: identity, Cov: covariance response, FD: finite-difference response (FD, $\tau\to0$). Rows with references are LiSM-adapted published solvers.
}
\label{tab:lism-ablation}
\endgroup
\end{table}

With an isotropic prior, the covariance response simply rescales the probe direction.
For a global probe on the mean token, the correction therefore shifts every patch by the same vector along that direction.
This uniform concept shift is consistent with the strong global adherence of Iso-Cov ($94.0$).
However, this same method fails on local tasks, with the lowest adherence (26.7).
In addition, for local tasks, the method presents high Hack: 72.2 for local vs. 5.9 for global (see Appendix~\ref{append:quant}).
This means that even though the method satisfies its own probe request in the representation, the image does not change.
These results suggest that isotropic scaling alone is insufficient for local tasks.

Adding a Jacobian response of the denoiser (JVP/VJP), or an approximation of it (FD), notably improves adherence on local tasks (Iso: 26.7\sarr71.4, Fit: 67.0\sarr83.0), at the cost of largely reduced adherence on global ones (Iso: 94.0\sarr50.0, Fit: 93.1\sarr61.3).
Similar local adherence is observed for the three responses (JVP, VJP, FD) when using the same weighting, as shown for Fit (83.0--84.6 local, but only 58.1--64.5 global).
In line with these quantitative results, using denoiser responses generally improves the quality of images for local tasks.
Without it, Fit-Id produces unrealistic images, such as zebras with multiple extra legs for keypoint requests (see Appendix~\ref{append:quali}).
Thus, using the denoiser response is crucial for local tasks. 

The Path response heavily mitigates the low global adherence of denoiser responses, improving this metric for all weightings over VJP (Point: $+6.1$ percentage points [pp], Iso: $+36.8$\,pp, Fit: $+32.3$\,pp).
For Point and Iso, however, this improvement comes at the cost of lower adherence in local tasks (87.4\sarr84.2 and 71.4\sarr63.2, respectively).
For Fit, on the other hand, local adherence barely changes (83.0--84.6$\sarr$83.2), while global adherence rises to $90.4$ from at most $64.5$ (FD).
Overall, GM adherence across tasks improves consistently with Path, with gains ranging from 1.7 pp (Point) to 16.6 pp (Fit).
The same holds for Hack, where Path variants present the lowest values by at least 7.9 pp.
We also noticed that under the same seed in local tasks, Fit-Path and Fit-FD often generate similar images, indicating that responses are nearly linear within the path range, something that does not happen for global tasks.

In general, while identity and covariance give the best global adherence and denoiser responses the best local adherence, neither succeeds on both.
Among responses, Path represents the most balanced alternative, with the highest GM adherence for every weighting
and the best overall result with Fit, where local adherence is barely affected.

\begin{figure}[t]
    \centering
    \begingroup
\definecolor{RAElism}{HTML}{3E6CB0}\definecolor{figgrey}{HTML}{A7ADB8}\definecolor{figink}{HTML}{222222}\definecolor{figmute}{HTML}{808080}\definecolor{figgrid}{HTML}{ECEDEF}\definecolor{figlegend}{HTML}{8A8F98}\definecolor{figframe}{HTML}{D0D3D8}
\def\figfs{\fontsize{6.8}{8.2}\selectfont}\def\figls{\fontsize{6.3}{7.6}\selectfont}\def\figqs{\fontsize{5.2}{6.2}\selectfont}
\edef\xunit{\the\dimexpr(\linewidth-27.77pt)/35\relax}
\edef\yunit{\the\dimexpr(\linewidth-27.77pt)/95\relax}
\noindent\begin{tikzpicture}[x=\xunit, y=\yunit, font=\footnotesize]
\useasboundingbox ([shift={(-23.77pt,-21.18pt)}]50,0) rectangle ([xshift=4.00pt]85,95);
\draw[figgrid, line width=0.5pt] (50,0) -- (50,95);
\draw[figgrid, line width=0.5pt] (55,0) -- (55,95);
\draw[figgrid, line width=0.5pt] (60,0) -- (60,95);
\draw[figgrid, line width=0.5pt] (65,0) -- (65,95);
\draw[figgrid, line width=0.5pt] (70,0) -- (70,95);
\draw[figgrid, line width=0.5pt] (75,0) -- (75,95);
\draw[figgrid, line width=0.5pt] (80,0) -- (80,95);
\draw[figgrid, line width=0.5pt] (85,0) -- (85,95);
\draw[figgrid, line width=0.5pt] (50,0) -- (85,0);
\draw[figgrid, line width=0.5pt] (50,20) -- (85,20);
\draw[figgrid, line width=0.5pt] (50,40) -- (85,40);
\draw[figgrid, line width=0.5pt] (50,60) -- (85,60);
\draw[figgrid, line width=0.5pt] (50,80) -- (85,80);
\draw[line width=0.6pt] (50,95) -- (50,0) -- (85,0);
\draw[line width=0.6pt] (50,0) -- ++(0,-2.5pt); \node[anchor=north, inner sep=0pt] at ([yshift=-4.5pt]50,0) {50};
\draw[line width=0.6pt] (55,0) -- ++(0,-2.5pt); \node[anchor=north, inner sep=0pt] at ([yshift=-4.5pt]55,0) {55};
\draw[line width=0.6pt] (60,0) -- ++(0,-2.5pt); \node[anchor=north, inner sep=0pt] at ([yshift=-4.5pt]60,0) {60};
\draw[line width=0.6pt] (65,0) -- ++(0,-2.5pt); \node[anchor=north, inner sep=0pt] at ([yshift=-4.5pt]65,0) {65};
\draw[line width=0.6pt] (70,0) -- ++(0,-2.5pt); \node[anchor=north, inner sep=0pt] at ([yshift=-4.5pt]70,0) {70};
\draw[line width=0.6pt] (75,0) -- ++(0,-2.5pt); \node[anchor=north, inner sep=0pt] at ([yshift=-4.5pt]75,0) {75};
\draw[line width=0.6pt] (80,0) -- ++(0,-2.5pt); \node[anchor=north, inner sep=0pt] at ([yshift=-4.5pt]80,0) {80};
\draw[line width=0.6pt] (85,0) -- ++(0,-2.5pt); \node[anchor=north, inner sep=0pt] at ([yshift=-4.5pt]85,0) {85};
\draw[line width=0.6pt] (50,0) -- ++(-2.5pt,0); \node[anchor=east, inner sep=0pt] at ([xshift=-4.5pt]50,0) {0};
\draw[line width=0.6pt] (50,20) -- ++(-2.5pt,0); \node[anchor=east, inner sep=0pt] at ([xshift=-4.5pt]50,20) {20};
\draw[line width=0.6pt] (50,40) -- ++(-2.5pt,0); \node[anchor=east, inner sep=0pt] at ([xshift=-4.5pt]50,40) {40};
\draw[line width=0.6pt] (50,60) -- ++(-2.5pt,0); \node[anchor=east, inner sep=0pt] at ([xshift=-4.5pt]50,60) {60};
\draw[line width=0.6pt] (50,80) -- ++(-2.5pt,0); \node[anchor=east, inner sep=0pt] at ([xshift=-4.5pt]50,80) {80};
\node[anchor=north, inner sep=0pt] at ([yshift=-13.91pt]67.5,0) {Preservation (GM) $\rightarrow$};
\node[anchor=south, rotate=90, inner sep=0pt] at ([xshift=-16.50pt]50,47.5) {Adherence (GM) $\rightarrow$};
\draw[RAElism, opacity=0.55, line width=0.9pt, dash pattern=on 2.70pt off 1.80pt] (72.77,87.21) -- (76.05,82.94) -- (78.53,50.50);
\node[circle, fill=figgrey, draw=white, line width=0.6pt, opacity=0.85, minimum size=5.98pt, inner sep=0pt] at (73.41,6.57) {};
\node[anchor=west, inner sep=0pt, text=figink, font=\figfs] at ([shift={(4.52pt,2.11pt)}]73.41,6.57) {DDNM};
\node[circle, fill=figgrey, draw=white, line width=0.6pt, opacity=0.85, minimum size=9.31pt, inner sep=0pt] at (71.83,82.71) {};
\node[anchor=east, inner sep=0pt, text=figink, font=\figfs] at ([shift={(-6.03pt,2.81pt)}]71.83,82.71) {DPS};
\node[circle, fill=figgrey, draw=white, line width=0.6pt, opacity=0.85, minimum size=9.31pt, inner sep=0pt] at (71.68,76.47) {};
\node[anchor=east, inner sep=0pt, text=figink, font=\figfs] at ([shift={(-6.03pt,2.81pt)}]71.68,76.47) {AdamDPS};
\node[circle, fill=figgrey, draw=white, line width=0.6pt, opacity=0.85, minimum size=5.89pt, inner sep=0pt] at (55.54,56.76) {};
\node[anchor=east, inner sep=0pt, text=figink, font=\figfs] at ([shift={(-4.48pt,2.09pt)}]55.54,56.76) {DDiff};
\node[circle, fill=figgrey, draw=white, line width=0.6pt, opacity=0.85, minimum size=15.86pt, inner sep=0pt] at (76.05,82.94) {};
\node[anchor=west, inner sep=0pt, text=figink, font=\figfs] at ([shift={(9.00pt,-4.20pt)}]76.05,82.94) {DCPS};
\node[circle, fill=figgrey, draw=white, line width=0.6pt, opacity=0.85, minimum size=5.90pt, inner sep=0pt] at (78.53,50.50) {};
\node[anchor=west, inner sep=0pt, text=figink, font=\figfs] at ([shift={(4.49pt,-2.09pt)}]78.53,50.50) {CoDPS};
\node[circle, fill=figgrey, draw=white, line width=0.6pt, opacity=0.85, minimum size=9.40pt, inner sep=0pt] at (69.73,61.61) {};
\node[anchor=east, inner sep=0pt, text=figink, font=\figfs] at ([shift={(-6.07pt,2.83pt)}]69.73,61.61) {$\Pi$GDM};
\node[circle, fill=figgrey, draw=white, line width=0.6pt, opacity=0.85, minimum size=11.00pt, inner sep=0pt] at (63.49,64.06) {};
\node[anchor=east, inner sep=0pt, text=figink, font=\figfs] at ([shift={(-6.80pt,3.17pt)}]63.49,64.06) {DTMPD};
\node[circle, fill=figgrey, draw=white, line width=0.6pt, opacity=0.85, minimum size=10.06pt, inner sep=0pt] at (71.95,51.01) {};
\node[anchor=west, align=left, inner sep=0pt, text=figink, font=\figfs] at ([shift={(6.37pt,-2.97pt)}]71.95,51.01) {Free\\Hunch};
\node[circle, fill=RAElism, draw=white, line width=0.6pt, minimum size=8.30pt, inner sep=0pt] at (72.77,87.21) {};
\node[anchor=east, inner sep=0pt, text=figink, font=\figfs] at ([shift={(-5.57pt,2.60pt)}]72.77,87.21) {\textbf{\textcolor{RAElism}{Fit-Path}}};
\node[circle, draw=figink, line width=0.9pt, minimum size=5.62pt, inner sep=0pt] at (81.39,5.08) {};
\node[anchor=east, inner sep=0pt, text=figink, font=\figfs] at ([shift={(-4.36pt,-2.03pt)}]81.39,5.08) {\textit{Control}};
\coordinate (lg) at (51.225,92.625);
\draw[figlegend, line width=0.7pt, dash pattern=on 2.10pt off 1.40pt, fill=white] (lg) rectangle ([shift={(37.29pt,-53.77pt)}]lg);
\node[font=\figfs, text=figink, inner sep=0pt] at ([shift={(18.65pt,-6.90pt)}]lg) {Time};
\node[circle, fill=figgrey, draw=white, line width=0.6pt, opacity=0.85, minimum size=6.00pt, inner sep=0pt] at ([shift={(11.48pt,-16.30pt)}]lg) {};
\node[anchor=base east, font=\figls, text=figink, inner sep=0pt] at ([shift={(29.77pt,-18.43pt)}]lg) {5};
\node[anchor=base west, font=\figls, text=figink, inner sep=0pt] at ([shift={(31.34pt,-18.43pt)}]lg) {s};
\node[circle, fill=figgrey, draw=white, line width=0.6pt, opacity=0.85, minimum size=9.00pt, inner sep=0pt] at ([shift={(11.48pt,-26.80pt)}]lg) {};
\node[anchor=base east, font=\figls, text=figink, inner sep=0pt] at ([shift={(29.77pt,-28.93pt)}]lg) {10};
\node[anchor=base west, font=\figls, text=figink, inner sep=0pt] at ([shift={(31.34pt,-28.93pt)}]lg) {s};
\node[circle, fill=figgrey, draw=white, line width=0.6pt, opacity=0.85, minimum size=15.97pt, inner sep=0pt] at ([shift={(11.48pt,-42.28pt)}]lg) {};
\node[anchor=base east, font=\figls, text=figink, inner sep=0pt] at ([shift={(29.77pt,-44.41pt)}]lg) {50};
\node[anchor=base west, font=\figls, text=figink, inner sep=0pt] at ([shift={(31.34pt,-44.41pt)}]lg) {s};
\node[anchor=south west, draw=figframe, line width=0.6pt, fill=white, inner sep=1.5pt] at (51.050,1.900) {\figfs\color{figink}\setlength{\tabcolsep}{3pt}\setlength{\arrayrulewidth}{0.5pt} \begin{tabular}{l|r|r} \textbf{Solver} & \textbf{GM} & \textbf{Time} \\  & Adh. {\tiny(Pres.)} & s$\,\downarrow$ \\ \noalign{\hrule height 0.9pt} \rowcolor{RAElism!15}\textcolor{RAElism}{\textbf{Fit-Path}} & \textcolor{RAElism}{\makebox[11.90pt][r]{\textbf{87.2}}\hspace{2.7pt}{\figqs (72.8)}} & \textcolor{RAElism}{8.5} \\ \hline DCPS~[\citenum{dcps}] & \makebox[11.90pt][r]{82.9}\hspace{2.7pt}{\figqs (76.1)} & 48.8 \\ DPS~[\citenum{dps}] & \makebox[11.90pt][r]{82.7}\hspace{2.7pt}{\figqs (71.8)} & 10.8 \\ AdamDPS~[\citenum{adamdps}] & \makebox[11.90pt][r]{76.5}\hspace{2.7pt}{\figqs (71.7)} & 10.7 \\ DTMPD~[\citenum{tmpd}] & \makebox[11.90pt][r]{64.1}\hspace{2.7pt}{\figqs (63.5)} & 15.9 \\ $\Pi$GDM~[\citenum{pigdm}] & \makebox[11.90pt][r]{61.6}\hspace{2.7pt}{\figqs (69.7)} & 11.0 \\ DDiff~[\citenum{ddiff}] & \makebox[11.90pt][r]{56.8}\hspace{2.7pt}{\figqs (55.5)} & 4.9 \\ Free Hunch~[\citenum{freehunch}] & \makebox[11.90pt][r]{51.0}\hspace{2.7pt}{\figqs (71.9)} & 12.8 \\ CoDPS~[\citenum{codps}] & \makebox[11.90pt][r]{50.5}\hspace{2.7pt}{\figqs (78.5)} & 4.9 \\ DDNM~[\citenum{ddnm}] & \makebox[11.90pt][r]{6.6}\hspace{2.7pt}{\figqs (73.4)} & 5.0 \\ \hline \textcolor{figmute}{\textit{Control}} & \textcolor{figmute}{\makebox[11.90pt][r]{\textit{5.1}}\hspace{2.7pt}{\figqs\itshape (81.4)}} & \textcolor{figmute}{\textit{4.6}} \\ \end{tabular}};
\end{tikzpicture}
\endgroup%
    \caption{\textbf{Comparison of Fit-Path with LiSM-adapted SOTA solvers.}
    Adherence (GM over the four tasks, $y$) versus preservation ($x$); marker area is time (s) per sample.
    The dashed blue line is the Pareto frontier. Control: same generator without guidance.}
    \label{fig:solver_sota}
\end{figure}

\subsection{Comparison with LiSM-adapted SOTA solvers}
Fig.~\ref{fig:solver_sota} compares state-of-the-art solvers adapted to LiSM with our best Path-based method, Fit-Path, using the geometric-mean (GM) adherence and preservation over the four tasks.
Fit-Path lies at the top end of the Pareto frontier over preservation and adherence, with the highest adherence (87.2) at $5.7\times$ less time than DCPS, the next-best on adherence, which provides slightly higher preservation (+3.3\,pp).
Interestingly, DDNM achieves marginally higher adherence than Control. 
A detailed analysis reveals that this method presents a very high Hack value (72.4), largely satisfying its own probe requests without updating the image accordingly.
Overall, these results support the use of Fit-Path for both global and local guidance in Semantic Mean-Space.

\subsection{Generalization to other VFMs}
Finally, we tested Fit-Path in the representation spaces of other VFMs, as is shown in Tab.~\ref{tab:rae-transfer}. Only the probes and covariance statistics are refit;
guidance strengths are reused from DINOv3 without retuning. LiSM transfers to both spaces successfully. Per-task adherence follows the encoder's linear readability: MAE is weak on class (47.5) but strong on keypoints (87.4), while SigLIP~2 shows the opposite pattern. LiSM can therefore benefit directly from more readable representation spaces from stronger VFMs.

\begin{table}[t]
\centering
\small
\begingroup
\definecolor{ctrlgray}{HTML}{808080}
\definecolor{RAElism}{HTML}{3E6CB0}
\setlength{\tabcolsep}{3.5pt}
\newcolumntype{z}[1]{>{\raggedleft\arraybackslash}p{#1pt}}
\begin{tabular}{l@{\hspace{6pt}}l|z{18}|z{25}z{25}z{25}z{25}}
\textbf{Encoder} & \textbf{Method} & \textbf{GM} & \textbf{Class} & \textbf{Attr.} & \textbf{Fg.} & \textbf{Kp.} \\
 & & & Acc. & CLIP & IoU & PCK \\
\noalign{\hrule height 0.8pt}
\textbf{MAE} & \textcolor{ctrlgray}{\textit{Control}} & \textcolor{ctrlgray}{0.0} & \textcolor{ctrlgray}{0.0} & \textcolor{ctrlgray}{16.7} & \textcolor{ctrlgray}{26.1} & \textcolor{ctrlgray}{16.5} \\
 & \textcolor{RAElism}{\textbf{LiSM}} & 73.3 & 47.5 & 91.9 & 75.8 & 87.4 \\
\hline
\textbf{SigLIP~2} & \textcolor{ctrlgray}{\textit{Control}} & \textcolor{ctrlgray}{0.0} & \textcolor{ctrlgray}{0.0} & \textcolor{ctrlgray}{15.2} & \textcolor{ctrlgray}{28.7} & \textcolor{ctrlgray}{12.9} \\
 & \textcolor{RAElism}{\textbf{LiSM}} & 82.6 & 79.0 & 95.7 & 77.2 & 79.5 \\
\hline
\textbf{DINOv3} & \textcolor{ctrlgray}{\textit{Control}} & \textcolor{ctrlgray}{0.0} & \textcolor{ctrlgray}{0.0} & \textcolor{ctrlgray}{15.7} & \textcolor{ctrlgray}{27.9} & \textcolor{ctrlgray}{16.8} \\
 & \textcolor{RAElism}{\textbf{LiSM}} & 86.7 & 89.0 & 91.9 & 77.2 & 89.5 \\
\end{tabular}
\caption{\textbf{Transfer of LiSM with Fit-Path solver across RAE encoders.}
Same ablation-subset requests. 
Metrics follow Tab.~\ref{tab:rae-main}.}
\label{tab:rae-transfer}
\endgroup
\end{table}

\section{Limitations}

LiSM inherits the readability of the encoder, so attributes that linear probes cannot read cannot be requested, and transfer across encoders follows probe accuracy.
Fitting probes and covariance requires labeled clean features.
The Gaussian weighting assumes a Gaussian prior on representations, which is a coarse model of a class-clustered space, and strong global requests reduce preservation.
Finally, while Fit-Path successfully balances performance in global and local requests, there is no single solver that is best for all tasks.
Closing these gaps represents valuable future work.

\section{Conclusion}

In this work, we proposed LiSM, a novel framework for training-free control of representation generators using linear probes.
Our work is founded on a simple but key observation: in representation flow models based on vision foundation models, such as RAE, a linear probe trained on clean data remains valid at every noise level, since it commutes with the posterior mean, so guidance becomes a Bayesian linear inverse problem.
With this, generation can be guided with any linear probe operating in the same representation space, such as a global class probe on the mean token or a per-patch probe for local control.

In our experiments, we have shown not only the overall suitability of the LiSM framework, but also its superior adherence over other state-of-the-art guidance frameworks such as pixel-space (e.g., DCPS) and noise-aware (e.g., CG) guidance.
Moreover, our systematic analysis of solvers showed that identity and covariance responses favor global tasks while denoiser responses favor local ones, and led to the proposal of Fit-Path, a LiSM solver that performs well on both with low probe hacking.

In conclusion, our framework offers a new and effective approach to lightweight and flexible control of representation generators.
Since LiSM relies only on frozen models and linear probes, we expect it to improve as vision foundation models and denoisers do.

\clearpage

\section*{Acknowledgment}
Funded by the European Union (ERC, HealthAEye, 101171183). Views and opinions expressed are however those of the author(s) only and do not necessarily reflect those of the European Union or the European Research Council. Neither the European Union nor the granting authority can be held responsible for them. This work was also supported in part by the Christian Doppler Research Association, Austrian Federal Ministry of Economy, Energy and Tourism, and the National Foundation for Research, Technology and Development.

{
    \small
    \bibliographystyle{ieeenat_fullname}
    \bibliography{lism}

\begin{thebibliography}{66}
\providecommand{\natexlab}[1]{#1}
\providecommand{\url}[1]{\texttt{#1}}
\expandafter\ifx\csname urlstyle\endcsname\relax
  \providecommand{\doi}[1]{doi: #1}\else
  \providecommand{\doi}{doi: \begingroup \urlstyle{rm}\Url}\fi

\bibitem[Bansal et~al.(2024)Bansal, Chu, Schwarzschild, Sengupta, Goldblum,
  Geiping, and Goldstein]{ug}
Arpit Bansal, Hong-Min Chu, Avi Schwarzschild, Roni Sengupta, Micah Goldblum,
  Jonas Geiping, and Tom Goldstein.
\newblock Universal guidance for diffusion models.
\newblock In \emph{ICLR}, 2024.

\bibitem[Belardi et~al.(2026)Belardi, Lovelace, Weinberger, and Gomes]{adamdps}
Christian Belardi, Justin Lovelace, Kilian Weinberger, and Carla Gomes.
\newblock Adaptive moments are surprisingly effective for plug-and-play
  diffusion sampling.
\newblock In \emph{ICLR}, 2026.

\bibitem[Bender and Morik(2026)]{didae}
Sidney Bender and Marco Morik.
\newblock Visual disentangled diffusion autoencoders: Scalable counterfactual
  generation for foundation models.
\newblock \emph{arXiv preprint arXiv:2601.21851}, 2026.

\bibitem[Boys et~al.(2024)Boys, Girolami, Pidstrigach, Reich, Mosca, and
  Akyildiz]{tmpd}
Benjamin Boys, Mark Girolami, Jakiw Pidstrigach, Sebastian Reich, Alan Mosca,
  and Omer~Deniz Akyildiz.
\newblock Tweedie moment projected diffusions for inverse problems.
\newblock \emph{TMLR}, 2024.

\bibitem[Chen et~al.(2023)Chen, Vi{\'e}gas, and Wattenberg]{beyondsurface}
Yida Chen, Fernanda Vi{\'e}gas, and Martin Wattenberg.
\newblock Beyond surface statistics: Scene representations in a latent
  diffusion model.
\newblock \emph{arXiv preprint arXiv:2306.05720}, 2023.

\bibitem[Chung et~al.(2023)Chung, Kim, McCann, Klasky, and Ye]{dps}
Hyungjin Chung, Jeongsol Kim, Michael~Thompson McCann, Marc~Louis Klasky, and
  Jong~Chul Ye.
\newblock Diffusion posterior sampling for general noisy inverse problems.
\newblock In \emph{ICLR}, 2023.

\bibitem[Deng et~al.(2009)Deng, Dong, Socher, Li, Li, and Fei-Fei]{imagenet}
Jia Deng, Wei Dong, Richard Socher, Li-Jia Li, Kai Li, and Li Fei-Fei.
\newblock {ImageNet}: A large-scale hierarchical image database.
\newblock In \emph{CVPR}, 2009.

\bibitem[Dhariwal and Nichol(2021)]{cg}
Prafulla Dhariwal and Alexander Nichol.
\newblock Diffusion models beat {GANs} on image synthesis.
\newblock In \emph{NeurIPS}, 2021.

\bibitem[Efron(2011)]{efron2011}
Bradley Efron.
\newblock Tweedie's formula and selection bias.
\newblock \emph{Journal of the American Statistical Association}, 106\penalty0
  (496), 2011.

\bibitem[Graikos et~al.(2025)Graikos, Jojic, and Samaras]{graikos2026fast}
Alexandros Graikos, Nebojsa Jojic, and Dimitris Samaras.
\newblock Fast constrained sampling in pre-trained diffusion models.
\newblock In \emph{NeurIPS}, 2025.

\bibitem[He et~al.(2022{\natexlab{a}})He, Yang, Yang, Kortylewski, Yuan, Chen,
  Liu, Yang, Yu, and Yuille]{partimagenet}
Ju He, Shuo Yang, Shaokang Yang, Adam Kortylewski, Xiaoding Yuan, Jie-Neng
  Chen, Shuai Liu, Cheng Yang, Qihang Yu, and Alan Yuille.
\newblock {PartImageNet}: A large, high-quality dataset of parts.
\newblock In \emph{ECCV}, 2022{\natexlab{a}}.

\bibitem[He et~al.(2016)He, Zhang, Ren, and Sun]{resnet}
Kaiming He, Xiangyu Zhang, Shaoqing Ren, and Jian Sun.
\newblock Deep residual learning for image recognition.
\newblock In \emph{CVPR}, 2016.

\bibitem[He et~al.(2022{\natexlab{b}})He, Chen, Xie, Li, Doll{\'a}r, and
  Girshick]{mae}
Kaiming He, Xinlei Chen, Saining Xie, Yanghao Li, Piotr Doll{\'a}r, and Ross
  Girshick.
\newblock Masked autoencoders are scalable vision learners.
\newblock In \emph{CVPR}, 2022{\natexlab{b}}.

\bibitem[Ho and Salimans(2022)]{cfg}
Jonathan Ho and Tim Salimans.
\newblock Classifier-free diffusion guidance.
\newblock \emph{arXiv preprint arXiv:2207.12598}, 2022.

\bibitem[Ho et~al.(2020)Ho, Jain, and Abbeel]{ddpm}
Jonathan Ho, Ajay Jain, and Pieter Abbeel.
\newblock Denoising diffusion probabilistic models.
\newblock In \emph{NeurIPS}, 2020.

\bibitem[Huben et~al.(2024)Huben, Cunningham, Smith, Ewart, and Sharkey]{sae}
Robert Huben, Hoagy Cunningham, Logan Smith, Aidan Ewart, and Lee Sharkey.
\newblock Sparse autoencoders find highly interpretable features in language
  models.
\newblock In \emph{ICLR}, 2024.

\bibitem[Janati et~al.(2024)Janati, Moufad, Durmus, Moulines, and Olsson]{dcps}
Yazid Janati, Badr Moufad, Alain Durmus, Eric Moulines, and Jimmy Olsson.
\newblock Divide-and-conquer posterior sampling for denoising diffusion priors.
\newblock In \emph{NeurIPS}, 2024.

\bibitem[Jose et~al.(2025)Jose, Moutakanni, Kang, Baldassarre, Darcet, Xu, Li,
  Szafraniec, Ramamonjisoa, Oquab, et~al.]{dinotxt}
Cijo Jose, Th{\'e}o Moutakanni, Dahyun Kang, Federico Baldassarre, Timoth{\'e}e
  Darcet, Hu Xu, Daniel Li, Marc Szafraniec, Micha{\"e}l Ramamonjisoa, Maxime
  Oquab, et~al.
\newblock Dinov2 meets text: A unified framework for image-and pixel-level
  vision-language alignment.
\newblock In \emph{CVPR}, 2025.

\bibitem[Kawar et~al.(2022)Kawar, Elad, Ermon, and Song]{ddrm}
Bahjat Kawar, Michael Elad, Stefano Ermon, and Jiaming Song.
\newblock Denoising diffusion restoration models.
\newblock In \emph{NeurIPS}, 2022.

\bibitem[Kim et~al.(2026)Kim, Levy, and Wetzstein]{ddiff}
Minseo Kim, Axel Levy, and Gordon Wetzstein.
\newblock Dual ascent diffusion for inverse problems.
\newblock In \emph{CVPR}, 2026.

\bibitem[Kouzelis et~al.(2025)Kouzelis, Karypidis, Kakogeorgiou, Gidaris, and
  Komodakis]{redi}
Theodoros Kouzelis, Efstathios Karypidis, Ioannis Kakogeorgiou, Spyridon
  Gidaris, and Nikos Komodakis.
\newblock Boosting generative image modeling via joint image-feature synthesis.
\newblock In \emph{NeurIPS}, 2025.

\bibitem[Kwon et~al.(2023)Kwon, Jeong, and Uh]{asyrp}
Mingi Kwon, Jaeseok Jeong, and Youngjung Uh.
\newblock Diffusion models already have a semantic latent space.
\newblock In \emph{ICLR}, 2023.

\bibitem[Kynk{\"a}{\"a}nniemi et~al.(2019)Kynk{\"a}{\"a}nniemi, Karras, Laine,
  Lehtinen, and Aila]{precision}
Tuomas Kynk{\"a}{\"a}nniemi, Tero Karras, Samuli Laine, Jaakko Lehtinen, and
  Timo Aila.
\newblock Improved precision and recall metric for assessing generative models.
\newblock In \emph{NeurIPS}, 2019.

\bibitem[Leng et~al.(2025)Leng, Singh, Hou, Xing, Xie, and Zheng]{repae}
Xingjian Leng, Jaskirat Singh, Yunzhong Hou, Zhenchang Xing, Saining Xie, and
  Liang Zheng.
\newblock {REPA-E}: Unlocking {VAE} for end-to-end tuning with latent diffusion
  transformers.
\newblock In \emph{ICCV}, 2025.

\bibitem[Li et~al.(2023{\natexlab{a}})Li, Patel, Vi{\'e}gas, Pfister, and
  Wattenberg]{iti}
Kenneth Li, Oam Patel, Fernanda Vi{\'e}gas, Hanspeter Pfister, and Martin
  Wattenberg.
\newblock Inference-time intervention: Eliciting truthful answers from a
  language model.
\newblock In \emph{NeurIPS}, 2023{\natexlab{a}}.

\bibitem[Li et~al.(2023{\natexlab{b}})Li, Liu, Wu, Mu, Yang, Gao, Li, and
  Lee]{gligen}
Yuheng Li, Haotian Liu, Qingyang Wu, Fangzhou Mu, Jianwei Yang, Jianfeng Gao,
  Chunyuan Li, and Yong~Jae Lee.
\newblock {GLIGEN}: Open-set grounded text-to-image generation.
\newblock In \emph{CVPR}, 2023{\natexlab{b}}.

\bibitem[Luo et~al.(2024)Luo, Darrell, Wang, Goldman, and
  Holynski]{readoutguidance}
Grace Luo, Trevor Darrell, Oliver Wang, Dan~B Goldman, and Aleksander Holynski.
\newblock Readout guidance: Learning control from diffusion features.
\newblock In \emph{CVPR}, 2024.

\bibitem[Manor and Michaeli(2024)]{ManorMichaeli}
Hila Manor and Tomer Michaeli.
\newblock On the posterior distribution in denoising: Application to
  uncertainty quantification.
\newblock In \emph{ICLR}, 2024.

\bibitem[Meng and Kabashima(2025)]{dmps}
Xiangming Meng and Yoshiyuki Kabashima.
\newblock Diffusion model based posterior sampling for noisy linear inverse
  problems.
\newblock In \emph{ACML}, pages 623--638. PMLR, 2025.

\bibitem[Moitra et~al.(2026)Moitra, Risteski, and Rohatgi]{moitra}
Ankur Moitra, Andrej Risteski, and Dhruv Rohatgi.
\newblock Steering diffusion models with quadratic rewards: a fine-grained
  analysis.
\newblock In \emph{COLT}, 2026.

\bibitem[Mou et~al.(2024)Mou, Wang, Xie, Wu, Zhang, Qi, and Shan]{t2iadapter}
Chong Mou, Xintao Wang, Liangbin Xie, Yanze Wu, Jian Zhang, Zhongang Qi, and
  Ying Shan.
\newblock T2i-adapter: Learning adapters to dig out more controllable ability
  for text-to-image diffusion models.
\newblock In \emph{AAAI}, 2024.

\bibitem[Nichol and Dhariwal(2021)]{iddpm}
Alexander~Quinn Nichol and Prafulla Dhariwal.
\newblock Improved denoising diffusion probabilistic models.
\newblock In \emph{ICML}, 2021.

\bibitem[Oquab et~al.(2024)Oquab, Darcet, Moutakanni, Vo, Szafraniec, Khalidov,
  Fernandez, Haziza, Massa, El-Nouby, Assran, Ballas, Galuba, Howes, Huang, Li,
  Misra, Rabbat, Sharma, Synnaeve, Xu, Jegou, Mairal, Labatut, Joulin, and
  Bojanowski]{dinov2}
Maxime Oquab, Timoth{\'e}e Darcet, Th{\'e}o Moutakanni, Huy~V. Vo, Marc
  Szafraniec, Vasil Khalidov, Pierre Fernandez, Daniel Haziza, Francisco Massa,
  Alaaeldin El-Nouby, Mido Assran, Nicolas Ballas, Wojciech Galuba, Russell
  Howes, Po-Yao Huang, Shang-Wen Li, Ishan Misra, Michael Rabbat, Vasu Sharma,
  Gabriel Synnaeve, Hu Xu, Herve Jegou, Julien Mairal, Patrick Labatut, Armand
  Joulin, and Piotr Bojanowski.
\newblock {DINO}v2: Learning robust visual features without supervision.
\newblock \emph{TMLR}, 2024.

\bibitem[Park et~al.(2024)Park, Choe, and Veitch]{linearrep}
Kiho Park, Yo~Joong Choe, and Victor Veitch.
\newblock The linear representation hypothesis and the geometry of large
  language models.
\newblock In \emph{ICML}, 2024.

\bibitem[Patterson and Hays(2012)]{sun}
Genevieve Patterson and James Hays.
\newblock {SUN} attribute database: Discovering, annotating, and recognizing
  scene attributes.
\newblock In \emph{CVPR}, 2012.

\bibitem[Qin et~al.(2022)Qin, Dai, Hu, Fan, Shao, and Van~Gool]{isnet}
Xuebin Qin, Hang Dai, Xiaobin Hu, Deng-Ping Fan, Ling Shao, and Luc Van~Gool.
\newblock Highly accurate dichotomous image segmentation.
\newblock In \emph{ECCV}, 2022.

\bibitem[Radford et~al.(2021)Radford, Kim, Hallacy, Ramesh, Goh, Agarwal,
  Sastry, Askell, Mishkin, Clark, et~al.]{clip}
Alec Radford, Jong~Wook Kim, Chris Hallacy, Aditya Ramesh, Gabriel Goh,
  Sandhini Agarwal, Girish Sastry, Amanda Askell, Pamela Mishkin, Jack Clark,
  et~al.
\newblock Learning transferable visual models from natural language
  supervision.
\newblock In \emph{ICML}, 2021.

\bibitem[Rissanen et~al.(2025)Rissanen, Heinonen, and Solin]{freehunch}
Severi Rissanen, Markus Heinonen, and Arno Solin.
\newblock Free hunch: Denoiser covariance estimation for diffusion models
  without extra costs.
\newblock In \emph{ICLR}, 2025.

\bibitem[Rombach et~al.(2022)Rombach, Blattmann, Lorenz, Esser, and Ommer]{ldm}
Robin Rombach, Andreas Blattmann, Dominik Lorenz, Patrick Esser, and Bj{\"o}rn
  Ommer.
\newblock High-resolution image synthesis with latent diffusion models.
\newblock In \emph{CVPR}, 2022.

\bibitem[Shen et~al.(2024)Shen, Jiang, Yang, Wang, Han, and
  Li]{shen2024understanding}
Yifei Shen, Xinyang Jiang, Yifan Yang, Yezhen Wang, Dongqi Han, and Dongsheng
  Li.
\newblock Understanding and improving training-free loss-based diffusion
  guidance.
\newblock In \emph{NeurIPS}, 2024.

\bibitem[Shi et~al.(2026)Shi, Wang, Zheng, Yuan, Wu, Wang, Wan, Zhou, and
  Lu]{svg}
Minglei Shi, Haolin Wang, Wenzhao Zheng, Ziyang Yuan, Xiaoshi Wu, Xintao Wang,
  Pengfei Wan, Jie Zhou, and Jiwen Lu.
\newblock Latent diffusion model without variational autoencoder.
\newblock In \emph{ICLR}, 2026.

\bibitem[Sim{\'e}oni et~al.(2026)Sim{\'e}oni, Vo, Seitzer, Baldassarre, Oquab,
  Jose, Khalidov, Szafraniec, Yi, Ramamonjisoa, Massa, Haziza, Wehrstedt, Wang,
  Darcet, Moutakanni, Sentana, Roberts, Vedaldi, Tolan, Brandt, Couprie,
  Mairal, Jegou, Labatut, and Bojanowski]{dinov3}
Oriane Sim{\'e}oni, Huy~V. Vo, Maximilian Seitzer, Federico Baldassarre, Maxime
  Oquab, Cijo Jose, Vasil Khalidov, Marc Szafraniec, Seung~Eun Yi, Michael
  Ramamonjisoa, Francisco Massa, Daniel Haziza, Luca Wehrstedt, Jianyuan Wang,
  Timoth{\'e}e Darcet, Th{\'e}o Moutakanni, Leonel Sentana, Claire Roberts,
  Andrea Vedaldi, Jamie Tolan, John Brandt, Camille Couprie, Julien Mairal,
  Herve Jegou, Patrick Labatut, and Piotr Bojanowski.
\newblock {DINO}v3.
\newblock \emph{TMLR}, 2026.

\bibitem[Singh et~al.(2026)Singh, Zheng, Wu, Zhang, Shechtman, and Xie]{raev2}
Jaskirat Singh, Boyang Zheng, Zongze Wu, Richard Zhang, Eli Shechtman, and
  Saining Xie.
\newblock Improved baselines with representation autoencoders.
\newblock In \emph{NeurIPS}, 2026.

\bibitem[Singhal et~al.(2025)Singhal, Horvitz, Teehan, Ren, Yu, McKeown, and
  Ranganath]{fksteering}
Raghav Singhal, Zachary Horvitz, Ryan Teehan, Mengye Ren, Zhou Yu, Kathleen
  McKeown, and Rajesh Ranganath.
\newblock A general framework for inference-time scaling and steering of
  diffusion models.
\newblock In \emph{ICML}, 2025.

\bibitem[Song et~al.(2023)Song, Vahdat, Mardani, and Kautz]{pigdm}
Jiaming Song, Arash Vahdat, Morteza Mardani, and Jan Kautz.
\newblock Pseudoinverse-guided diffusion models for inverse problems.
\newblock In \emph{ICLR}, 2023.

\bibitem[Sun et~al.(2019)Sun, Xiao, Liu, and Wang]{hrnet}
Ke Sun, Bin Xiao, Dong Liu, and Jingdong Wang.
\newblock Deep high-resolution representation learning for human pose
  estimation.
\newblock In \emph{CVPR}, 2019.

\bibitem[Surkov et~al.(2025)Surkov, Wendler, Mari, Terekhov, Deschenaux, West,
  Gulcehre, and Bau]{sdxlsae}
Viacheslav Surkov, Chris Wendler, Antonio Mari, Mikhail Terekhov, Justin
  Deschenaux, Robert West, Caglar Gulcehre, and David Bau.
\newblock One-step is enough: Sparse autoencoders for text-to-image diffusion
  models.
\newblock In \emph{NeurIPS}, 2025.

\bibitem[Tong et~al.(2026)Tong, Zheng, Wang, Tang, Ma, Brown, Yang, Fergus,
  LeCun, and Xie]{scalerae}
Shengbang Tong, Boyang Zheng, Ziteng Wang, Bingda Tang, Nanye Ma, Ellis Brown,
  Jihan Yang, Rob Fergus, Yann LeCun, and Saining Xie.
\newblock Scaling text-to-image diffusion transformers with representation
  autoencoders.
\newblock \emph{arXiv preprint arXiv:2601.16208}, 2026.

\bibitem[Touvron et~al.(2022)Touvron, Cord, and J{\'e}gou]{deitiii}
Hugo Touvron, Matthieu Cord, and Herv{\'e} J{\'e}gou.
\newblock Deit iii: Revenge of the vit.
\newblock In \emph{ECCV}, 2022.

\bibitem[Tschannen et~al.(2025)Tschannen, Gritsenko, Wang, Naeem,
  Alabdulmohsin, Parthasarathy, Evans, Beyer, Xia, Mustafa, et~al.]{siglip2}
Michael Tschannen, Alexey Gritsenko, Xiao Wang, Muhammad~Ferjad Naeem, Ibrahim
  Alabdulmohsin, Nikhil Parthasarathy, Talfan Evans, Lucas Beyer, Ye Xia, Basil
  Mustafa, et~al.
\newblock {SigLIP} 2: Multilingual vision-language encoders with improved
  semantic understanding, localization, and dense features.
\newblock \emph{arXiv preprint arXiv:2502.14786}, 2025.

\bibitem[Wang et~al.(2023)Wang, Yu, and Zhang]{ddnm}
Yinhuai Wang, Jiwen Yu, and Jian Zhang.
\newblock Zero-shot image restoration using denoising diffusion null-space
  model.
\newblock In \emph{ICLR}, 2023.

\bibitem[Wu et~al.(2025)Wu, Zhang, Shi, Gao, Chen, Wang, Chen, Gao, Tang,
  Cheng, et~al.]{reg}
Ge Wu, Shen Zhang, Ruijing Shi, Shanghua Gao, Zhenyuan Chen, Lei Wang, Zhaowei
  Chen, Hongcheng Gao, Yao Tang, Ming-Ming Cheng, et~al.
\newblock Representation entanglement for generation: Training diffusion
  transformers is much easier than you think.
\newblock In \emph{NeurIPS}, 2025.

\bibitem[Wu et~al.(2023)Wu, Trippe, Naesseth, Blei, and Cunningham]{tds}
Luhuan Wu, Brian Trippe, Christian Naesseth, David Blei, and John Cunningham.
\newblock Practical and asymptotically exact conditional sampling in diffusion
  models.
\newblock In \emph{NeurIPS}, 2023.

\bibitem[Xiao et~al.(2018)Xiao, Wu, and Wei]{resnet_keypoints}
Bin Xiao, Haiping Wu, and Yichen Wei.
\newblock Simple baselines for human pose estimation and tracking.
\newblock In \emph{ECCV}, 2018.

\bibitem[Yao et~al.(2025)Yao, Yang, and Wang]{vavae}
Jingfeng Yao, Bin Yang, and Xinggang Wang.
\newblock Reconstruction vs. generation: Taming optimization dilemma in latent
  diffusion models.
\newblock In \emph{CVPR}, 2025.

\bibitem[Ye et~al.(2024)Ye, Lin, Han, Xu, Liu, Liang, Ma, Zou, and Ermon]{tfg}
Haotian Ye, Haowei Lin, Jiaqi Han, Minkai Xu, Sheng Liu, Yitao Liang, Jianzhu
  Ma, James Zou, and Stefano Ermon.
\newblock {TFG}: Unified training-free guidance for diffusion models.
\newblock In \emph{NeurIPS}, 2024.

\bibitem[Yismaw et~al.(2025)Yismaw, Kamilov, and Asif]{codps}
Nebiyou Yismaw, Ulugbek~S Kamilov, and M~Salman Asif.
\newblock Gaussian is all you need: A unified framework for solving inverse
  problems via diffusion posterior sampling.
\newblock \emph{IEEE Transactions on Computational Imaging}, 2025.

\bibitem[Yu et~al.(2021)Yu, Xu, Zhang, Zhao, Guan, and Tao]{ap10k}
Hang Yu, Yufei Xu, Jing Zhang, Wei Zhao, Ziyu Guan, and Dacheng Tao.
\newblock {AP-10K: A Benchmark for Animal Pose Estimation in the Wild}.
\newblock In \emph{NeurIPS Systems Track on Datasets and Benchmarks}, 2021.

\bibitem[Yu et~al.(2023)Yu, Wang, Zhao, Ghanem, and Zhang]{freedom}
Jiwen Yu, Yinhuai Wang, Chen Zhao, Bernard Ghanem, and Jian Zhang.
\newblock {FreeDoM}: Training-free energy-guided conditional diffusion model.
\newblock In \emph{ICCV}, 2023.

\bibitem[Yu et~al.(2025)Yu, Kwak, Jang, Jeong, Huang, Shin, and Xie]{repa}
Sihyun Yu, Sangkyung Kwak, Huiwon Jang, Jongheon Jeong, Jonathan Huang, Jinwoo
  Shin, and Saining Xie.
\newblock Representation alignment for generation: Training diffusion
  transformers is easier than you think.
\newblock In \emph{ICLR}, 2025.

\bibitem[Zhai et~al.(2023)Zhai, Mustafa, Kolesnikov, and Beyer]{siglip}
Xiaohua Zhai, Basil Mustafa, Alexander Kolesnikov, and Lucas Beyer.
\newblock Sigmoid loss for language image pre-training.
\newblock In \emph{ICCV}, 2023.

\bibitem[Zhang et~al.(2023)Zhang, Rao, and Agrawala]{controlnet}
Lvmin Zhang, Anyi Rao, and Maneesh Agrawala.
\newblock Adding conditional control to text-to-image diffusion models.
\newblock In \emph{ICCV}, 2023.

\bibitem[Zheng et~al.(2026)Zheng, Ma, Tong, and Xie]{rae}
Boyang Zheng, Nanye Ma, Shengbang Tong, and Saining Xie.
\newblock Diffusion transformers with representation autoencoders.
\newblock In \emph{ICLR}, 2026.

\bibitem[Zheng et~al.(2024)Zheng, Gao, Fan, Liu, Laaksonen, Ouyang, and
  Sebe]{birefnet}
Peng Zheng, Dehong Gao, Deng-Ping Fan, Li Liu, Jorma Laaksonen, Wanli Ouyang,
  and Nicu Sebe.
\newblock Bilateral reference for high-resolution dichotomous image
  segmentation.
\newblock \emph{CAAI Artificial Intelligence Research}, 3, 2024.

\bibitem[Zhou et~al.(2017)Zhou, Lapedriza, Khosla, Oliva, and
  Torralba]{places365}
Bolei Zhou, Agata Lapedriza, Aditya Khosla, Aude Oliva, and Antonio Torralba.
\newblock Places: A 10 million image database for scene recognition.
\newblock \emph{IEEE TPAMI}, 40\penalty0 (6), 2017.

\bibitem[Zhu et~al.(2023)Zhu, Zhang, Liang, Cao, Wen, Timofte, and
  Van~Gool]{diffpir}
Yuanzhi Zhu, Kai Zhang, Jingyun Liang, Jiezhang Cao, Bihan Wen, Radu Timofte,
  and Luc Van~Gool.
\newblock Denoising diffusion models for plug-and-play image restoration.
\newblock In \emph{CVPRW}, 2023.

\end{thebibliography}
}
\clearpage
\appendix
\finalhyphendemerits=10000 \widowpenalty=10000 \clubpenalty=10000 \parfillskip=0pt plus 0.8\linewidth

\section{LiSM Request and Solvers Detail}
\label{append:solver-detail}

\subsection{Requests}
All methods share the same overall request design, and most use the same loss. For a few existing solvers that fail with this default design, we tune the request specifically to improve their results (see ``Solver-specific requests''). No LiSM ablation variant is tuned in this way.
In our experiments, all requirements are hard constraints ($\beta_i=0$ in Eq.~(\ref{eq:likelihood})), so for every task the weight function reduces to
\begin{equation}
L(y)=\prod_{i=1}^{q}\mathbf 1\{y_i\ge 0\},
\end{equation}
and a request is satisfied only when all its scores are nonnegative. The tasks differ in the number of scores $q$ and in how each request score $y_i$ (Sec.~\ref{sec:requests}) is built from probe logits.

\paragraph{Class.}
For a requested class $k$, the index $i$ runs over the 999 rival classes $j\neq k$, and each contributes the margin between their logits,
\begin{equation}
y_i=s_k-s_j,
\end{equation}
so the request holds when the target logit is at least as large as every rival logit.\looseness=-1

\paragraph{Attributes.}
An attribute request has a single score ($q=1$), the logit of the requested attribute $a$ on the spatially pooled representation,
\begin{equation}
y_1=s_a.
\end{equation}
The other six attributes are left free.

\paragraph{Foreground.}
A foreground request specifies a mask $M\in\{0,1\}^{256}$ on the $16\times16$ patch grid, and $i$ runs over the 256 patches. With $s_i$ the foreground probe logit at patch $i$,
\begin{equation}
y_i=(2M_i-1)\,(s_i-\rho),\qquad \rho=\mathrm{logit}(0.2),
\end{equation}
so foreground patches must reach a foreground probability of $0.2$, and background patches must not exceed it.

\paragraph{Keypoints.}
A keypoint request places each visible joint $v\in V$ at a patch $p_v$. Let $R_v$ be the set of all patches outside the $3\times3$ neighborhood of $p_v$; the index $i$ runs over all pairs $(v,p')$ with $p'\in R_v$. With $s_{v,p}$ the joint's probe logit at patch $p$, each pair requires a unit margin,
\begin{equation}
y_i=s_{v,p_v}-s_{v,p'}-1.
\end{equation}

\paragraph{Solver-specific requests.} Free Hunch~\cite{freehunch} uses Gaussian equality targets. For class guidance, setting all 999 margins to \(1\) gives the target class a softmax probability of only \(0.0027\), even when every equality holds. We therefore target a margin of \(8\) over the current strongest rival. For attributes, Free Hunch targets logit \(1\); for foreground and background, it targets logits \(\rho+1\) and \(\rho-1\); for each visible keypoint, it targets a margin of \(1\) over the strongest non-neighbor rival. Pixel-space DDiff~\cite{ddiff} keeps the class request but uses the cross-entropy \(-\log p_k\), where \(p_k\) is the classifier's softmax probability of class \(k\), as the squared probability residual gave poor class control in pilot experiments. DDNM~\cite{ddnm} uses a minimum-norm correction on the violated constraints. It requires a class margin of \(1\) and uses \(\rho+1\) and \(\rho-1\) as one-sided foreground and background bounds. Its attribute and keypoint constraints follow the defaults.

\subsection{Existing Solvers Adaptation}
\paragraph{DDNM.} DDNM~\cite{ddnm} projects a clean prediction onto a linear measurement equality. We adapt this step to semantic inequalities.
Write the requested bounds as \(Az+c\geq 0\), and let \(\mathcal B=\{i:(A\hat z+c)_i<0\}\) index the bounds violated by the current clean prediction. We apply one joint pseudoinverse correction:
\[
\hat z_{\mathrm{DDNM}}
=\hat z+A_{\mathcal B}^\dagger\bigl[-(A_{\mathcal B}\hat z+c_{\mathcal B})\bigr].
\]
Only violated rows contribute to the correction. We recompute \(\mathcal B\) at each denoising step.
A single correction can change which bounds are satisfied, so this is one step toward their intersection.

\paragraph{Other solvers.} DPS~\cite{dps} differentiates the task loss through the readout and the clean prediction. CoDPS~\cite{codps} and \(\Pi\)GDM~\cite{pigdm} use isotropic uncertainty in the request scores, while DTMPD~\cite{tmpd} estimates their variances from denoiser Jacobian products. Free Hunch~\cite{freehunch} uses the Gaussian observation targets defined above and carries its online covariance estimate across steps.
Some solvers also retain optimizer state or perform inner inference. AdamDPS~\cite{adamdps} applies its bias-corrected gradient update to the task loss and carries its first and second moments across steps. DDiff applies the task loss in its measurement update and carries its dual variable into the next reverse step. DCPS~\cite{dcps} evaluates the semantic loss on a clean estimate within each observation block, runs Langevin updates on entering the block, and samples the next state from its fitted variational proposal.\looseness=-1

\subsection{Fitting Gaussian Covariance}
For class and attribute requests, we fit the covariance of the spatially pooled clean features. For foreground and keypoints, we retain the full spatial grid, e.g. $16\times16\times1024$ representation and apply randomized PCA directly to the centered training features. We retain 16 principal directions and model the remaining variance with an isotropic residual. For foreground, we additionally estimate a residual variance for the probe score at each patch. This factorization lets us compute the covariance needed by each request without forming a dense $(16\times16\times1024)^2$ matrix.

The fitted covariance $\hat\Sigma_0$ of clean representations enters through its posterior given the noisy state (Eq.~(\ref{eq:cov-transfer})),
\begin{equation}
\hat\Sigma_t=\bigl(\hat\Sigma_0^{-1}+k_t I\bigr)^{-1},\qquad k_t=\Bigl(\tfrac{1-t}{t}\Bigr)^{2},
\end{equation}
which keeps the eigenvectors of $\hat\Sigma_0$ and maps each eigenvalue $\sigma^2$ to $\sigma^2/(1+k_t\sigma^2)$.
For pooled features, averaging $P$ noisy patches gives $\sigma^2/(1+Pk_t\sigma^2)$.

\subsection{PyTorch-style pseudocode}
\label{append:torch}
\begin{algorithm}[t]
\caption{~LiSM sampling with Fit-Id or Fit-Path using an Euler ODE integrator, in PyTorch-like style.}
\label{alg:lism-torch}
\begin{lstlisting}[style=torch]
# f: denoiser, predicts clean tokens
# z: [P, D] tokens, t: 1 -> 0
# global: W [Q, D], b [Q] signed probe;
#         V [D, D], S [D] pooled covariance
# local:  w [D], b probe;  y [P] = +1 / -1 mask;
#         V [P, K, D], S [K], s residual

for t, t_next in zip(times, times[1:]):
    # (1) clean prediction
    z_hat = f(z, t)
    k = ((1 - t) / t) ** 2

    # (1) mean scores: mu = A z_hat + c
    if request.kind == "global":
        mu = W @ z_hat.mean(0) + b
        S_t = S / (1 + P * k * S)
        var = (W @ V) ** 2 @ S_t
    elif request.kind == "local":
        mu = y * (z_hat @ w + b)
        S_t, s_t = S / (1 + k * S), s / (1 + k * s)
        var = (V @ w) ** 2 @ (S_t - s_t) + s_t * (w @ w)

    # (2) request weights: lam = G(L, mu), Fit
    sd = var.sqrt()
    lam = pdf(mu / sd) / cdf(mu / sd) / sd
    # (2) probe direction: u = A^T lam
    if request.kind == "global":
        u = W.T @ lam / P
    else:
        u = (lam * y)[:, None] * w

    # (3) representation correction: R(u)
    if response == "Identity":
        R = u
    elif response == "Path":
        c_t = t ** 2 / (1 - t)
        size = u.norm(dim=-1).max()
        shift = c_t * size
        r = min(shift, H)
        z_hat_q = f(z + r * u / size, t)
        R = shift / r * (z_hat_q - z_hat)
    delta = alpha * R

    # (4) guided velocity
    v = (z - (z_hat + delta)) / t
    # Euler step
    z = z + (t_next - t) * v
\end{lstlisting}
\end{algorithm}

Alg.~\ref{alg:lism-torch} gives PyTorch-style pseudocode of Alg.~\ref{alg:lism} for two of the best-performing LiSM solvers, Fit-Id and Fit-Path.\looseness=-1

\section{Additional Experiments}
\label{append:exp}
\subsection{Linearity of the Denoiser Response}
In Tab.~\ref{tab:lism-ablation}, the Path response and the local responses (FD,
JVP, VJP) reach similar adherence on local tasks, while Path is clearly better
on global ones. Here we examine where this difference comes from.

Fit-Path and Fit-FD share the request weights and the probe direction $u$, but they critically differ in the query radius: $H{=}2$ for Path and $H{=}0.2$ for FD,
which approximates the JVP. If the denoiser response
$s \mapsto f_\theta(z_t + s u, t)$ were linear along $u$, the two corrections
would coincide at every step. Any difference between them must therefore come
from a nonlinear response within the path. We hypothesize that this
nonlinearity is specific to global requests, and test where it occurs.

\paragraph{Setup.}
We run unguided trajectories on the development requests. At each step, we
compute the Fit probe direction $u$ as if guidance were applied at that state,
evaluate the response $f_\theta(z_t + \tau u, t) - \hat z$ at the two query
radii, and measure the cosine similarity between the two. Since each solver
only rescales its response, this is also the cosine similarity between the
Fit-Path and Fit-FD corrections $\Delta$. Using a shared unguided trajectory
ensures that both are compared at the same state.\looseness=-1

\paragraph{Results.}
Tab.~\ref{tab:supp-nonlinear} reports medians at four noise levels. For
foreground and keypoints, the two responses agree almost perfectly at every
noise level (cosine $\geq 0.99$). This matches the uncurated samples in
Figs.~\ref{fig:supp_quali_foreground} and~\ref{fig:supp_quali_keypoints}, where Fit-FD, Fit-JVP, Fit-VJP
and Fit-Path produce nearly identical images from the same initial noise.
For class and attributes, the responses also agree for $t \leq 0.81$, but
disagree strongly at high noise: the cosine similarity is $0.48$ for class and
$0.68$ for attributes at $t{=}0.995$, and $0.33$ and $0.52$ at $t{=}0.95$.

In this nonlinear regime, a local response extrapolates from a neighbourhood
that does not represent the path, while Path measures the response over the
displacement that is actually requested. Since nonlinearity is the only way
the two can differ, this accounts for the gap between Path and local responses
on global tasks, and for their agreement on local ones.\looseness=-1

\begin{table}[t]
\centering
\small
\begingroup
\definecolor{RAElism}{HTML}{3E6CB0}
\setlength{\tabcolsep}{3.5pt}
\newcolumntype{z}[1]{>{\raggedleft\arraybackslash}p{#1pt}}
\begin{tabular}{l|z{25}z{25}z{25}z{25}|z{30}}
\textbf{Task} & \multicolumn{4}{c|}{\textbf{$\cos(\Delta_{\text{FD}},\Delta_{\text{Path}})$}} & \textbf{Linear} \\
Noise $t$ & 0.995 & 0.95 & 0.81 & 0.59 & for $t$ \\
\noalign{\hrule height 0.8pt}
Class & \cellcolor{RAElism!12}0.48 & \cellcolor{RAElism!12}0.33 & 0.98 & 1.00 & $\le$0.81 \\
Attr. & \cellcolor{RAElism!12}0.68 & \cellcolor{RAElism!12}0.52 & 0.98 & 1.00 & $\le$0.81 \\
\hline
Fg. & 1.00 & 1.00 & 1.00 & 1.00 & all \\
Kp. & 1.00 & 0.99 & 1.00 & 1.00 & all \\
\end{tabular}
\caption{\textbf{Where the denoiser response is nonlinear.}
Cosine similarity between the Fit-FD and Fit-Path corrections, $\cos(\Delta_{\mathrm{FD}},\Delta_{\mathrm{Path}})$, which query the denoiser along the request direction $u$ with a small ($H{=}0.2$) or a large ($H{=}2$) step. Medians over unguided trajectories at four noise levels $t$.
A value of 1.00 means both corrections point the same way, so the response is linear along $u$; shaded cells (below 0.95) mark a nonlinear response.
Linear for $t$: noise levels at which the response is linear.}
\label{tab:supp-nonlinear}
\endgroup
\end{table}

\subsection{Additional applications of LiSM}
\paragraph{Composition control.}
\begin{table}[t]
\centering
\small
\begingroup
\definecolor{ctrlgray}{HTML}{808080}
\setlength{\tabcolsep}{3.5pt}
\newcolumntype{z}[1]{>{\raggedleft\arraybackslash}p{#1pt}}
\begin{tabular}{l|z{22}z{22}z{22}z{22}}
\textbf{Requests} & \textbf{Class} & \textbf{Attr.} & \textbf{PCK} & \textbf{All} \\
\noalign{\hrule height 0.8pt}
Class + attr. & 66.8 & 84.5 & -- & 54.8 \\
Class + attr. + kpts. & 82.3 & 90.9 & 85.7 & 58.4 \\
\hline
\textcolor{ctrlgray}{Cond. + attr.} & \textcolor{ctrlgray}{56.9} & \textcolor{ctrlgray}{92.7} & \textcolor{ctrlgray}{--} & \textcolor{ctrlgray}{--} \\
\textcolor{ctrlgray}{Cond. + kpts.} & \textcolor{ctrlgray}{88.0} & \textcolor{ctrlgray}{--} & \textcolor{ctrlgray}{90.0} & \textcolor{ctrlgray}{--} \\
\end{tabular}
\caption{\textbf{Composing requests without class conditioning.}
Black rows: Fit-Path under the null class label, with the class as one of the guided requests, on the attribute test set (first row) and the keypoint test set (second row).
Grey rows (Cond.): for reference, single-task Fit-Path under class conditioning on the attribute and the keypoint test set.
All: images that meet every request at once, with the pose counted as met at a PCK of at least 0.8.}
\label{tab:supp-composition}
\endgroup
\end{table}

\begin{figure}
    \centering
    \includegraphics[width=1\linewidth]{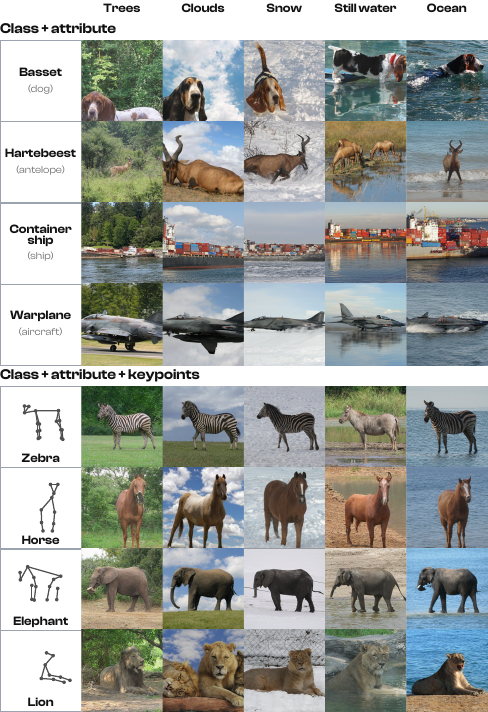}
    \caption{\textbf{Qualitative results of guidance composition by LiSM.} We show that it is possible to guide both class and attributes, or class, attributes, and keypoints together, with LiSM using the Fit-Path solver.}
    \label{fig:supp-comp}
\end{figure}
In the main experiments, the denoiser is conditioned on the requested class
for the attribute, foreground and keypoint tasks, so that all methods are
compared on the guided task alone. LiSM can also supply the class itself. We
run the denoiser under the null class and guide several requests at once with
Fit-Path: class with an attribute, and class with an attribute and keypoints.
Guidance strengths are selected on the development set.

Tab.~\ref{tab:supp-composition} compares these with single-task Fit-Path
under class conditioning. With class and attribute guided together, class
recognition is 10\,pp higher and attribute adherence 8\,pp lower. Adding
keypoints as a third request lowers class recognition and PCK by 4--6\,pp
relative to keypoint-only guidance, while the attribute is satisfied in 91\%
of the images. Fig.~\ref{fig:supp-comp} shows qualitative examples.
Overall, it can be observed that LiSM effectively supports composition of different requests.

\paragraph{Text control.}
\begin{figure}
    \centering
    \includegraphics[width=1\linewidth]{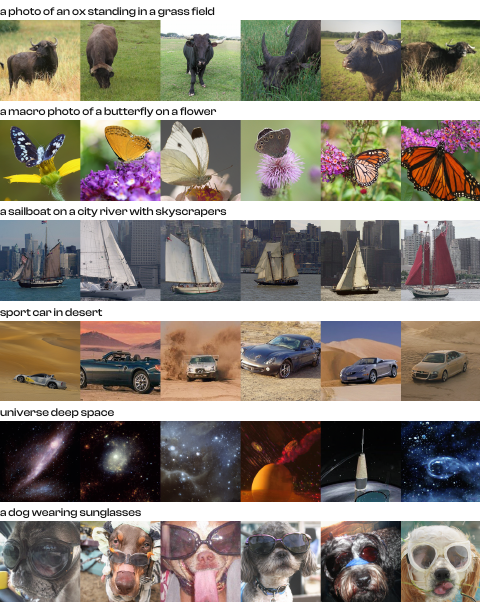}
    \caption{\textbf{Qualitative results of zero-shot text-to-image generation with LiSM.} Using text, it is possible to directly guide the object, attributes, and background at the same time. Notably, we found that it can also generate realistic new combinations, such as a dog wearing sunglasses.}
    \label{fig:supp-t2i}
\end{figure}
LiSM can also take a text prompt as request. DINOv3 provides a text encoder $E_\mathrm{L}$ aligned with an image encoder $E_\mathrm{V}$, trained with the \texttt{dino.txt} method~\cite{dinotxt}. $E_\mathrm{V}$ is the same frozen DINOv3 backbone as our RAE encoder $E$ with a light vision head on top, so the two representations are closely related. Therefore, we simply bridge them by fitting a single linear layer on 50{,}000 ImageNet images without captions, that maps the pooled representation of $E$, used for class and attributes guidance, to the embedding of $E_\mathrm{V}$, which is naturally also the space of $E_\mathrm{L}$. The similarity to the prompt embedding $E_\mathrm{L}(\text{prompt})$ is then a linear score on the representation, and we request that it exceeds a fixed threshold, using Fit-Path as before.

We show some qualitative results in Fig.~\ref{fig:supp-t2i}. Notably, with this approach we can not only achieve easy composition control for foreground and background, but also generate new composition, such as ``sports car in desert" or ``a dog wearing sunglasses".\looseness=-1

\section{Implementation Details}
\begin{table}[t]
\centering
\small
\begingroup
\setlength{\tabcolsep}{3pt}
\newcolumntype{z}[1]{>{\raggedleft\arraybackslash}p{#1pt}}
\newcommand{\support}[1]{{\scriptsize #1}}
\begin{tabular}{@{}lll|z{26}z{26}@{}}
\textbf{Task} & \textbf{Family} & \textbf{Predictor} & \textbf{Target} & \textbf{Agree.} \\
\noalign{\hrule height 0.8pt}
\textbf{Class}                                & LiSM        & Linear                                   & 78.7 & 80.1 \\
\support{ImageNet~\cite{imagenet}}   & Noise-aware & MLP                                      & 78.6 & 80.0 \\
                                     & Pixel-space & ResNet-50~\cite{resnet}                  & 88.5 & 91.1 \\
                                     & Evaluation  & DeiT-III-L~\cite{deitiii}                & 93.7 & --   \\
\hline
\textbf{Attributes}                           & LiSM        & Linear                                   & 69.4 & 92.5 \\
\support{SUN Attr.~\cite{sun}}       & Noise-aware & MLP                                      & 69.3 & 92.6 \\
                                     & Pixel-space & Places365~\cite{places365}        & 66.9 & 91.2 \\
                                     & Evaluation  & CLIP~\cite{clip}                         & 63.6 & --   \\
\hline
\textbf{Foreground}                           & LiSM        & Linear                                   & 69.5 & 77.9 \\
\support{PartImageNet~\cite{partimagenet}} & Noise-aware & CNN                                & 77.0 & 86.7 \\
                                     & Pixel-space & IS-Net~\cite{isnet}                      & 78.4 & 87.5 \\
                                     & Evaluation  & BiRefNet~\cite{birefnet}                 & 82.9 & --   \\
\hline
\textbf{Keypoints}                            & LiSM        & Linear                                   & 90.5 & 88.4 \\
\support{AP-10K~\cite{ap10k}}        & Noise-aware & CNN                                      & 95.9 & 92.6 \\
                                     & Pixel-space & ResNet-50~\cite{resnet,resnet_keypoints} & 91.4 & 93.6 \\
                                     & Evaluation  & HRNet~\cite{hrnet}                       & 92.6 & --   \\
\end{tabular}
\caption{\textbf{Task predictors and support datasets.}
Target: performance against the dataset target; Agree.: agreement with the external evaluator; both on a 0--100 scale.
Noise-aware predictors are evaluated at $t=0$.}
\label{tab:task_predictors}
\endgroup
\end{table}

\label{append:imple-detail}

\paragraph{Sampling and guidance.}
For the four task experiments we use the pretrained ImageNet RAEv2 model~\cite{raev2} with a DINOv3-L encoder, reconstruction decoder and DiT-DH-XL denoiser. Image resolution is $256\times256$, with patch size $16\times16$, forming a $16\times16$ grid. The encoder, decoder, and denoiser remain frozen throughout our experiments. We follow the default RAEv2 sampling grid of 100 steps with time-shift factor 8. The base flow sampler uses Euler integration. We disable RAEv2's internal guidance. For class guidance, the denoiser receives the null-class context so that the guidance method alone supplies the requested class. For attribute, foreground, and keypoint guidance, we condition the denoiser on the requested ImageNet class. Solvers are not guided at the first sampling step, where $t=1$ and the correction scale $c_t=t^2/(1-t)$ is undefined. $c_t$ remains very large in the steps that follow, so on our 100-step grid this leaves the first four steps unguided. For keypoints, the solvers are guided from the second step. All methods use the same initial Gaussian noise for each request–seed pair.

For each task, solvers within a guidance framework share the same predictor: a linear probe, a noise-aware predictor, or a pixel-space network. On each task’s support dataset, Table~\ref{tab:task_predictors} reports both how well each predictor matches the annotations and how closely it agrees with the independent evaluator used to score generated images. We evaluate the noise-aware predictors at the clean endpoint $t=0$.

\paragraph{Metrics.}
We evaluate generated images using two metrics, adherence and preservation, computed (after decoding) using the external networks shown in Table~\ref{tab:task_predictors}.

Adherence measures the fulfillment of the request. For class, we use the DeiT top-1 accuracy on the requested class. For attributes, we fit linear readouts for a CLIP vision encoder to align with attribute score, one linear layer per attribute. Then, we report the fraction of images whose score for the requested attribute exceeds 0.1. For foreground, we average the probability outputs of BiRefNet~\cite{birefnet} within each patch and mark a cell as foreground if its mean exceeds \(0.2\). We report mean intersection over union (IoU) with the requested mask on the $16\times16$ grid. For keypoints, we report the percentage of correct keypoints (PCK@0.2). A visible joint is correct if HRNet~\cite{hrnet} predicts it within $20\%$ of the longer side of the target bounding box with a confidence of at least $0.2$. We average over visible joints within each image, and then over images.

Preservation measures if the generated image is still a valid sample. For class guidance, we use precision~\cite{precision}: the fraction of generated images whose features are considered within the reference distribution, which we calculate from a support set of 46{,}000 images from ImageNet. For the other three tasks, we use the class top-1 accuracy as preservation.

\paragraph{Tuning the guidance strength.}
For each solver with a tunable guidance strength, we sweep candidate values separately for each task on a fixed development set. For class guidance, we select the strength with the highest DeiT top-1 accuracy on the requested class. For the other three tasks, we select the highest task adherence among candidates that have at least 75\% class preservation. The selection scores are the fitted CLIP score for attributes, mask IoU for foreground, and PCK@0.2 for keypoints. We fix the selected strengths before test evaluation. Experiments were run on NVIDIA A100 and GeForce RTX 5080 GPUs. Runtime was measured in a separate run on the former device.

\paragraph{Request sets.}
The development set contains 100 class, 84 attribute (12 per attribute), 100 foreground, and 100 keypoint request--seed pairs, 384 in total.
The test set is ten times larger in every task: 1{,}000 class, 840 attribute (120 per attribute), 1{,}000 foreground, and 1{,}000 keypoint pairs, 3{,}840 in total.
Ablations use a fixed subset of 810 test pairs (200 class, 210 attribute, 200 foreground, and 200 keypoint).
We design the sets to be diverse, spanning requests that the class-conditional generator already tends to satisfy as well as ones that conflict with its prior.
All requests are drawn at random before
any image is generated and are never filtered by model outputs. Each request is
repeated over several seeds, and all methods share the same requests and
initial noise.

For \emph{Class}, we request every ImageNet class, with the generator under the null class label. The development set uses 100 randomly drawn classes, and the ablation subset uses 200 classes drawn at random from the remaining 900. For \emph{Attributes}, we use seven scene attributes from SUN Attributes~\cite{sun} and pair each of them with
randomly drawn ImageNet classes that span animals and vehicles. We do not
require a pair to co-occur in real images, so the set contains both common
pairs, such as a hartebeest on grass, and unusual ones, such as a motor scooter
on the ocean.
For \emph{Foreground}, we randomly draw ImageNet classes from PartImageNet~\cite{partimagenet} and use the annotated object masks of real images as requests. Every class receives a mask from an image of the same class (\emph{within-class}), which the generator can plausibly produce, and a mask transferred from another class. The transferred mask comes either from a class of the same super-category (\emph{within-category}), such as another quadruped, or from a different super-category (\emph{cross-category}), which asks for a shape that the class rarely takes.
For \emph{Keypoints}, we use animal species that are in both AP-10K~\cite{ap10k} and ImageNet, and draw real
poses from AP-10K as requests. Every pose is requested once with its own
species (\emph{within-species}) and once with another species
(\emph{cross-species}), without adapting the pose to the new body proportions.\looseness=-1

\section{Full Quantitative Results}
\label{append:quant} 
Full per task results for Tab.~\ref{tab:rae-main}, Tab.~\ref{tab:lism-ablation} and Fig.~\ref{fig:solver_sota} are shown in Tab.~\ref{tab:supp-test-full}, Tab.~\ref{tab:supp-ablation-full} and Tab.~\ref{tab:supp-solver-full} respectively.

\begin{table*}[t]
\centering
\small
\begingroup
\definecolor{ctrlgray}{HTML}{808080}
\definecolor{RAEpixel}{HTML}{566274}
\definecolor{RAEnoise}{HTML}{8A5675}
\definecolor{RAElism}{HTML}{3E6CB0}
\setlength{\tabcolsep}{3pt}
\newcolumntype{z}[1]{>{\raggedleft\arraybackslash}p{#1pt}}
\begin{tabular}{l|z{24}z{24}||*{5}{z{24}}|*{5}{z{24}}}
\textbf{Method} & \multicolumn{2}{r||}{\textbf{GM}} & \multicolumn{5}{c|}{\textbf{Class}} & \multicolumn{5}{c}{\textbf{Attributes}} \\
 & \multicolumn{2}{r||}{Adh. {\scriptsize (Pres.)}} & Adh. & Int. & Hack & Pres. & Time & Adh. & Int. & Hack & Pres. & Time \\
\noalign{\hrule height 0.8pt}
\textcolor{ctrlgray}{\textit{Ref. (CFG)}} & \multicolumn{2}{r||}{\textcolor{ctrlgray}{--}} & \textcolor{ctrlgray}{84.0} & \textcolor{ctrlgray}{--} & \textcolor{ctrlgray}{--} & \textcolor{ctrlgray}{73.9} & \textcolor{ctrlgray}{4.6} & \textcolor{ctrlgray}{--} & \textcolor{ctrlgray}{--} & \textcolor{ctrlgray}{--} & \textcolor{ctrlgray}{--} & \textcolor{ctrlgray}{--} \\
\hline
\textcolor{ctrlgray}{\textit{Control}} & \multicolumn{2}{r||}{\textcolor{ctrlgray}{\mbox{5.1\,{\scriptsize (81.4)}}}} & \textcolor{ctrlgray}{0.1} & \textcolor{ctrlgray}{--} & \textcolor{ctrlgray}{--} & \textcolor{ctrlgray}{62.1} & \textcolor{ctrlgray}{4.5} & \textcolor{ctrlgray}{15.2} & \textcolor{ctrlgray}{--} & \textcolor{ctrlgray}{--} & \textcolor{ctrlgray}{85.8} & \textcolor{ctrlgray}{4.6} \\
\hline
\rowcolor{RAEpixel!12}
\multicolumn{13}{l}{\textcolor{RAEpixel}{\textbf{\textit{Pixel-space}}}} \\
UG\,\cite{ug} & \multicolumn{2}{r||}{\mbox{32.4\,{\scriptsize (53.8)}}} & 18.5 & 94.2 & 75.7 & 16.2 & 56.4 & 27.1 & 28.7 & 1.5 & 80.8 & 54.3 \\
DPS\,\cite{dps} & \multicolumn{2}{r||}{\mbox{28.0\,{\scriptsize (61.7)}}} & 21.2 & 45.8 & 24.6 & 25.0 & 17.8 & 19.8 & 35.6 & 15.8 & 80.7 & 17.1 \\
AdamDPS\,\cite{adamdps} & \multicolumn{2}{r||}{\mbox{21.4\,{\scriptsize (51.1)}}} & 10.7 & 89.3 & 78.6 & 13.0 & 17.8 & 19.6 & 13.1 & 0.0 & 81.4 & 17.1 \\
DDiff\,\cite{ddiff} & \multicolumn{2}{r||}{\mbox{38.8\,{\scriptsize (55.5)}}} & 31.0 & 73.8 & 42.8 & 14.9 & 11.9 & 35.8 & 99.5 & 63.7 & 84.8 & 11.2 \\
DCPS\,\cite{dcps} & \multicolumn{2}{r||}{\mbox{18.9\,{\scriptsize (76.4)}}} & 1.8 & 28.2 & 26.4 & 56.3 & 65.7 & 34.3 & 62.6 & 28.3 & 73.1 & 63.4 \\
\hline
\rowcolor{RAEnoise!12}
\multicolumn{13}{l}{\textcolor{RAEnoise}{\textbf{\textit{Noise-aware}}}} \\
CG\,\cite{cg} & \multicolumn{2}{r||}{\mbox{72.1\,{\scriptsize (78.0)}}} & \underline{92.2} & 99.8 & 7.6 & 64.7 & 4.9 & 72.9 & 85.2 & 12.4 & 77.6 & 4.9 \\
AdamCG\,\cite{adamdps} & \multicolumn{2}{r||}{\mbox{65.0\,{\scriptsize (74.0)}}} & 87.9 & 99.6 & 11.7 & 48.7 & 4.9 & 63.3 & 79.8 & 16.4 & 81.7 & 4.9 \\
\hline
\rowcolor{RAElism!12}
\multicolumn{13}{l}{\textcolor{RAElism}{\textbf{\textit{LiSM (ours)}}}} \\
Fit-Id & \multicolumn{2}{r||}{\mbox{\underline{79.5}\,{\scriptsize (76.0)}}} & \textbf{93.6} & 100.0 & 6.4 & 67.8 & 4.9 & \textbf{94.5} & 100.0 & 5.5 & 64.4 & 4.9 \\
\textbf{Fit-Path} & \multicolumn{2}{r||}{\mbox{\textbf{87.2}\,{\scriptsize (72.8)}}} & 90.2 & 94.4 & 4.2 & 67.8 & 6.7 & \underline{92.7} & 96.9 & 4.2 & 56.9 & 7.6 \\
\multicolumn{13}{c}{} \\
\textbf{Method} & \textbf{Hack} & \textbf{Time} & \multicolumn{5}{c|}{\textbf{Foreground}} & \multicolumn{5}{c}{\textbf{Keypoints}} \\
 & $\downarrow$ & s\,$\downarrow$ & Adh. & Int. & Hack & Pres. & Time & Adh. & Int. & Hack & Pres. & Time \\
\noalign{\hrule height 0.8pt}
\textcolor{ctrlgray}{\textit{Control}} & \textcolor{ctrlgray}{--} & \textcolor{ctrlgray}{4.6} & \textcolor{ctrlgray}{27.4} & \textcolor{ctrlgray}{--} & \textcolor{ctrlgray}{--} & \textcolor{ctrlgray}{87.6} & \textcolor{ctrlgray}{4.6} & \textcolor{ctrlgray}{16.0} & \textcolor{ctrlgray}{--} & \textcolor{ctrlgray}{--} & \textcolor{ctrlgray}{94.0} & \textcolor{ctrlgray}{4.6} \\
\hline
\rowcolor{RAEpixel!12}
\multicolumn{13}{l}{\textcolor{RAEpixel}{\textbf{\textit{Pixel-space}}}} \\
UG\,\cite{ug} & 26.9 & 70.6 & 47.0 & 60.4 & 13.4 & 79.8 & 111.4 & 46.5 & 63.3 & 16.8 & 79.9 & 60.2 \\
DPS\,\cite{dps} & 18.1 & 22.5 & 38.4 & 53.1 & 14.7 & 82.2 & 35.6 & 38.4 & 55.8 & 17.3 & 87.5 & 19.4 \\
AdamDPS\,\cite{adamdps} & 28.1 & 22.5 & 31.6 & 35.4 & 3.8 & 86.8 & 35.6 & 31.6 & 61.6 & 30.0 & 74.2 & 19.4 \\
DDiff\,\cite{ddiff} & 31.6 & 16.5 & 38.6 & 48.9 & 10.3 & 84.4 & 29.6 & 52.7 & 62.3 & 9.5 & 89.0 & 13.4 \\
DCPS\,\cite{dcps} & 16.8 & 76.6 & \underline{66.4} & 78.8 & 12.4 & 86.3 & 107.3 & 31.4 & 31.4 & 0.0 & 95.8 & 70.1 \\
\hline
\rowcolor{RAEnoise!12}
\multicolumn{13}{l}{\textcolor{RAEnoise}{\textbf{\textit{Noise-aware}}}} \\
CG\,\cite{cg} & 7.2 & 5.0 & 44.9 & 47.3 & 2.4 & 85.0 & 5.0 & \underline{89.8} & 96.1 & 6.3 & 86.7 & 5.0 \\
AdamCG\,\cite{adamdps} & 9.5 & 5.0 & 38.0 & 39.4 & 1.4 & 85.5 & 5.0 & 84.2 & 92.6 & 8.4 & 88.2 & 5.1 \\
\hline
\rowcolor{RAElism!12}
\multicolumn{13}{l}{\textcolor{RAElism}{\textbf{\textit{LiSM (ours)}}}} \\
Fit-Id & 19.1 & 5.1 & 58.8 & 100.0 & 41.2 & 82.5 & 5.0 & 76.9 & 100.0 & 23.1 & 92.4 & 5.4 \\
\textbf{Fit-Path} & 4.7 & 8.5 & \textbf{76.8} & 78.8 & 2.0 & 82.6 & 9.6 & \textbf{90.0} & 98.5 & 8.5 & 88.0 & 10.2 \\
\end{tabular}
\caption{\textbf{Extended version of Tab.~\ref{tab:rae-main}: guidance performance on the test set.}
GM: geometric mean of adherence (Adh.) and of preservation (Pres.) over the four tasks; the Hack and Time columns below it are means over the four tasks.
The other columns give the same results per task: adherence, internal adherence (Int.), Hack, preservation, and time in seconds per image.
Control: generator without guidance toward the scored task (null for class; class-conditioned otherwise).
\textbf{Bold}: best; \underline{underline}: second-best.}
\label{tab:supp-test-full}
\endgroup
\end{table*}

\begin{table*}[t]
\centering
\small
\begingroup
\setlength{\tabcolsep}{3pt}
\newcolumntype{z}[1]{>{\raggedleft\arraybackslash}p{#1pt}}
\begin{tabular}{ll|z{24}z{24}|z{44}||*{5}{z{24}}|*{5}{z{24}}}
$\mathcal G$ & $\mathcal R$ & \multicolumn{2}{r|}{\textbf{GM}} & \textbf{Global} & \multicolumn{5}{c|}{\textbf{Class}} & \multicolumn{5}{c}{\textbf{Attributes}} \\
 & & \multicolumn{2}{r|}{Adh. {\scriptsize (Pres.)}} & Adh. {\scriptsize (Pres.)} & Adh. & Int. & Hack & Pres. & Time & Adh. & Int. & Hack & Pres. & Time \\
\noalign{\hrule height 0.8pt}
Point & VJP\,\cite{dps} & \multicolumn{2}{r|}{\mbox{81.0\,{\scriptsize (74.5)}}} & \mbox{75.1\,{\scriptsize (63.9)}} & 73.5 & 99.0 & 25.5 & 67.0 & 10.8 & 76.7 & 95.7 & 19.0 & 61.0 & 10.7 \\
Point & Path & \multicolumn{2}{r|}{\mbox{\underline{82.7}\,{\scriptsize (76.2)}}} & \mbox{81.2\,{\scriptsize (72.3)}} & 90.0 & 95.0 & 5.0 & 66.5 & 9.6 & 73.3 & 63.8 & 0.0 & 78.6 & 9.7 \\
\hline
Iso & Cov\,\cite{codps} & \multicolumn{2}{r|}{\mbox{50.1\,{\scriptsize (77.5)}}} & \mbox{\textbf{94.0}\,{\scriptsize (68.1)}} & 91.5 & 100.0 & 8.5 & 75.5 & 4.7 & 96.7 & 100.0 & 3.3 & 61.4 & 4.7 \\
Iso & VJP\,\cite{pigdm} & \multicolumn{2}{r|}{\mbox{59.7\,{\scriptsize (73.6)}}} & \mbox{50.0\,{\scriptsize (64.9)}} & 46.0 & 84.5 & 38.5 & 52.0 & 10.5 & 54.3 & 44.3 & 0.0 & 81.0 & 10.5 \\
Iso & Path & \multicolumn{2}{r|}{\mbox{74.1\,{\scriptsize (75.4)}}} & \mbox{86.8\,{\scriptsize (65.9)}} & 90.0 & 95.0 & 5.0 & 66.5 & 6.6 & 83.8 & 74.8 & 0.0 & 65.2 & 8.9 \\
\hline
Fit & Id & \multicolumn{2}{r|}{\mbox{79.0\,{\scriptsize (74.6)}}} & \mbox{\underline{93.1}\,{\scriptsize (65.4)}} & 92.0 & 100.0 & 8.0 & 67.0 & 4.9 & 94.3 & 100.0 & 5.7 & 63.8 & 4.9 \\
Fit & JVP & \multicolumn{2}{r|}{\mbox{71.3\,{\scriptsize (69.8)}}} & \mbox{61.3\,{\scriptsize (58.9)}} & 58.5 & 94.0 & 35.5 & 54.0 & 28.5 & 64.3 & 100.0 & 35.7 & 64.3 & 33.1 \\
Fit & FD & \multicolumn{2}{r|}{\mbox{73.2\,{\scriptsize (67.6)}}} & \mbox{64.5\,{\scriptsize (55.0)}} & 57.5 & 95.0 & 37.5 & 50.5 & 8.3 & 72.4 & 100.0 & 27.6 & 60.0 & 8.6 \\
Fit & VJP & \multicolumn{2}{r|}{\mbox{70.1\,{\scriptsize (69.3)}}} & \mbox{58.1\,{\scriptsize (58.6)}} & 55.0 & 91.5 & 36.5 & 50.0 & 10.3 & 61.4 & 100.0 & 38.6 & 68.6 & 10.6 \\
Fit & Path & \multicolumn{2}{r|}{\mbox{\textbf{86.7}\,{\scriptsize (70.9)}}} & \mbox{90.4\,{\scriptsize (61.1)}} & 89.0 & 92.5 & 3.5 & 67.0 & 6.7 & 91.9 & 97.1 & 5.2 & 55.7 & 7.6 \\
\multicolumn{15}{c}{} \\
$\mathcal G$ & $\mathcal R$ & \textbf{Hack} & \textbf{Time} & \textbf{Local} & \multicolumn{5}{c|}{\textbf{Foreground}} & \multicolumn{5}{c}{\textbf{Keypoints}} \\
 & & $\downarrow$ & s\,$\downarrow$ & Adh. {\scriptsize (Pres.)} & Adh. & Int. & Hack & Pres. & Time & Adh. & Int. & Hack & Pres. & Time \\
\noalign{\hrule height 0.8pt}
Point & VJP\,\cite{dps} & 13.7 & 10.8 & \mbox{\textbf{87.4}\,{\scriptsize (86.9)}} & 79.9 & 87.2 & 7.3 & 91.0 & 10.8 & 95.6 & 98.6 & 3.0 & 83.0 & 10.7 \\
Point & Path & 2.8 & 9.8 & \mbox{84.2\,{\scriptsize (80.2)}} & 79.6 & 83.4 & 3.7 & 81.5 & 9.7 & 89.1 & 91.5 & 2.4 & 79.0 & 10.3 \\
\hline
Iso & Cov\,\cite{codps} & 39.1 & 4.9 & \mbox{26.7\,{\scriptsize (88.1)}} & 34.9 & 100.0 & 65.1 & 93.5 & 4.9 & 20.5 & 99.8 & 79.4 & 83.0 & 5.3 \\
Iso & VJP\,\cite{pigdm} & 11.0 & 11.0 & \mbox{71.4\,{\scriptsize (83.4)}} & 54.0 & 50.1 & 0.0 & 87.5 & 11.3 & 94.5 & 99.8 & 5.3 & 79.5 & 11.7 \\
Iso & Path & 3.1 & 9.0 & \mbox{63.2\,{\scriptsize (86.2)}} & 43.4 & 39.5 & 0.0 & 87.5 & 9.7 & 92.1 & 99.6 & 7.5 & 85.0 & 10.7 \\
\hline
Fit & Id & 19.7 & 5.1 & \mbox{67.0\,{\scriptsize (85.1)}} & 60.2 & 100.0 & 39.8 & 90.0 & 5.0 & 74.7 & 100.0 & 25.3 & 80.5 & 5.4 \\
Fit & JVP & 20.4 & 34.6 & \mbox{83.0\,{\scriptsize (82.7)}} & 77.0 & 78.1 & 1.2 & 84.0 & 37.6 & 89.4 & 98.8 & 9.4 & 81.5 & 39.2 \\
Fit & FD & 18.9 & 9.1 & \mbox{83.0\,{\scriptsize (83.0)}} & 76.9 & 78.2 & 1.3 & 84.0 & 9.6 & 89.4 & 98.7 & 9.2 & 82.0 & 10.2 \\
Fit & VJP & 21.3 & 11.0 & \mbox{\underline{84.6}\,{\scriptsize (81.9)}} & 78.7 & 81.0 & 2.2 & 85.0 & 11.2 & 90.9 & 98.9 & 8.0 & 79.0 & 11.8 \\
Fit & Path & 4.8 & 8.5 & \mbox{83.2\,{\scriptsize (82.2)}} & 77.2 & 78.3 & 1.1 & 84.5 & 9.6 & 89.5 & 98.7 & 9.2 & 80.0 & 10.2 \\
\end{tabular}
\caption{\textbf{Extended version of Tab.~\ref{tab:lism-ablation}: ablation of weighting $\mathcal G$ and response $\mathcal R$ on the ablation subset.}
The first columns repeat the main table: GM, Global and Local are geometric means of adherence (Adh.) and of preservation (Pres.) over all four tasks, over class and attributes, and over foreground and keypoints.
The Hack and Time columns beside Local are means over all four tasks.
The other columns give the same results per task: adherence, internal adherence (Int.), Hack, preservation, and time in seconds per image.
\textbf{Bold}: best; \underline{underline}: second-best.}
\label{tab:supp-ablation-full}
\endgroup
\end{table*}

\begin{table*}[t]
\centering
\small
\begingroup
\definecolor{ctrlgray}{HTML}{808080}
\definecolor{RAEpixel}{HTML}{566274}
\definecolor{RAEnoise}{HTML}{8A5675}
\definecolor{RAElism}{HTML}{3E6CB0}
\setlength{\tabcolsep}{3pt}
\newcolumntype{z}[1]{>{\raggedleft\arraybackslash}p{#1pt}}
\doublerulesepcolor{white}
\begin{tabular}{l|z{24}z{24}||*{5}{z{24}}|*{5}{z{24}}}
\textbf{Solver} & \multicolumn{2}{r||}{\textbf{GM}} & \multicolumn{5}{c|}{\textbf{Class}} & \multicolumn{5}{c}{\textbf{Attributes}} \\
 & \multicolumn{2}{r||}{Adh. {\scriptsize (Pres.)}} & Adh. & Int. & Hack & Pres. & Time & Adh. & Int. & Hack & Pres. & Time \\
\noalign{\hrule height 0.8pt}
\doublerulesepcolor{RAElism!12}
\rowcolor{RAElism!12}
\textbf{Fit-Path} & \multicolumn{2}{r||}{\mbox{\textbf{87.2}\,{\scriptsize (72.8)}}} & 90.2 & 94.4 & 4.2 & 67.8 & 6.7 & 92.7 & 96.9 & 4.2 & 56.9 & 7.6 \\
\doublerulesepcolor{white}
\hline
DCPS\,\cite{dcps} & \multicolumn{2}{r||}{\mbox{82.9\,{\scriptsize (76.1)}}} & 78.9 & 98.7 & 19.8 & 71.9 & 49.0 & 75.7 & 85.0 & 9.3 & 62.7 & 48.7 \\
DPS\,\cite{dps} & \multicolumn{2}{r||}{\mbox{82.7\,{\scriptsize (71.8)}}} & 79.2 & 99.1 & 19.9 & 61.6 & 10.8 & 78.0 & 96.2 & 18.2 & 61.2 & 10.7 \\
AdamDPS\,\cite{adamdps} & \multicolumn{2}{r||}{\mbox{76.5\,{\scriptsize (71.7)}}} & 76.3 & 99.6 & 23.3 & 56.1 & 10.8 & 63.0 & 81.7 & 18.7 & 72.7 & 10.7 \\
DTMPD\,\cite{tmpd} & \multicolumn{2}{r||}{\mbox{64.1\,{\scriptsize (63.5)}}} & 68.1 & 93.4 & 25.3 & 30.8 & 15.8 & 65.1 & 100.0 & 34.9 & 70.1 & 15.7 \\
$\Pi$GDM\,\cite{pigdm} & \multicolumn{2}{r||}{\mbox{61.6\,{\scriptsize (69.7)}}} & 49.8 & 85.2 & 35.4 & 52.5 & 10.5 & 54.3 & 50.4 & 0.0 & 75.4 & 10.5 \\
DDiff\,\cite{ddiff} & \multicolumn{2}{r||}{\mbox{56.8\,{\scriptsize (55.5)}}} & 55.8 & 100.0 & 44.2 & 18.0 & 5.0 & 70.1 & 100.0 & 29.9 & 77.1 & 4.7 \\
Free Hunch\,\cite{freehunch} & \multicolumn{2}{r||}{\mbox{51.0\,{\scriptsize (71.9)}}} & 90.8 & 100.0 & 9.2 & 70.4 & 12.8 & 57.0 & 99.8 & 42.7 & 50.1 & 12.7 \\
CoDPS\,\cite{codps} & \multicolumn{2}{r||}{\mbox{50.5\,{\scriptsize (78.5)}}} & 93.4 & 100.0 & 6.6 & 73.0 & 4.7 & 96.0 & 100.0 & 4.0 & 64.3 & 4.7 \\
DDNM\,\cite{ddnm} & \multicolumn{2}{r||}{\mbox{6.6\,{\scriptsize (73.4)}}} & 0.2 & 100.0 & 99.8 & 42.0 & 5.0 & 21.1 & 55.1 & 34.0 & 83.6 & 4.9 \\
\hline
\textcolor{ctrlgray}{\textit{Control}} & \multicolumn{2}{r||}{\textcolor{ctrlgray}{\mbox{5.1\,{\scriptsize (81.4)}}}} & \textcolor{ctrlgray}{0.1} & \textcolor{ctrlgray}{--} & \textcolor{ctrlgray}{--} & \textcolor{ctrlgray}{62.1} & \textcolor{ctrlgray}{4.5} & \textcolor{ctrlgray}{15.2} & \textcolor{ctrlgray}{--} & \textcolor{ctrlgray}{--} & \textcolor{ctrlgray}{85.8} & \textcolor{ctrlgray}{4.6} \\
\multicolumn{13}{c}{} \\
\textbf{Solver} & \textbf{Hack} & \textbf{Time} & \multicolumn{5}{c|}{\textbf{Foreground}} & \multicolumn{5}{c}{\textbf{Keypoints}} \\
 & $\downarrow$ & s\,$\downarrow$ & Adh. & Int. & Hack & Pres. & Time & Adh. & Int. & Hack & Pres. & Time \\
\noalign{\hrule height 0.8pt}
\doublerulesepcolor{RAElism!12}
\rowcolor{RAElism!12}
\textbf{Fit-Path} & 4.7 & \multicolumn{1}{z{24}||}{8.5} & 76.8 & 78.8 & 2.0 & 82.6 & 9.6 & 90.0 & 98.5 & 8.5 & 88.0 & 10.2 \\
\doublerulesepcolor{white}
\hline
DCPS\,\cite{dcps} & 8.2 & 48.8 & 83.8 & 87.6 & 3.7 & 81.5 & 48.9 & 94.5 & 94.5 & 0.1 & 91.0 & 48.7 \\
DPS\,\cite{dps} & 12.1 & 10.8 & 79.6 & 87.0 & 7.4 & 85.9 & 10.8 & 95.2 & 98.2 & 3.0 & 82.2 & 10.7 \\
AdamDPS\,\cite{adamdps} & 12.9 & 10.7 & 75.9 & 81.9 & 6.0 & 82.4 & 10.8 & 93.7 & 97.2 & 3.4 & 78.5 & 10.7 \\
DTMPD\,\cite{tmpd} & 19.8 & 15.9 & 47.8 & 46.7 & 0.0 & 83.3 & 15.8 & 79.4 & 98.2 & 18.8 & 90.3 & 16.3 \\
$\Pi$GDM\,\cite{pigdm} & 10.0 & 11.0 & 56.0 & 51.2 & 0.0 & 83.0 & 11.3 & 95.2 & 99.7 & 4.6 & 72.0 & 11.7 \\
DDiff\,\cite{ddiff} & 29.7 & 4.9 & 47.6 & 50.7 & 3.1 & 74.1 & 5.0 & 55.7 & 97.6 & 41.9 & 92.5 & 4.7 \\
Free Hunch\,\cite{freehunch} & 21.5 & 12.8 & 30.4 & 30.1 & 0.0 & 86.5 & 12.8 & 43.0 & 77.1 & 34.1 & 87.8 & 12.8 \\
CoDPS\,\cite{codps} & 38.8 & 4.9 & 34.2 & 100.0 & 65.8 & 87.8 & 4.9 & 21.2 & 99.9 & 78.6 & 92.3 & 5.3 \\
DDNM\,\cite{ddnm} & 72.4 & 5.0 & 27.5 & 100.0 & 72.5 & 88.2 & 4.9 & 16.1 & 99.4 & 83.3 & 93.8 & 5.1 \\
\hline
\textcolor{ctrlgray}{\textit{Control}} & \textcolor{ctrlgray}{--} & \textcolor{ctrlgray}{4.6} & \textcolor{ctrlgray}{27.4} & \textcolor{ctrlgray}{--} & \textcolor{ctrlgray}{--} & \textcolor{ctrlgray}{87.6} & \textcolor{ctrlgray}{4.6} & \textcolor{ctrlgray}{16.0} & \textcolor{ctrlgray}{--} & \textcolor{ctrlgray}{--} & \textcolor{ctrlgray}{94.0} & \textcolor{ctrlgray}{4.6} \\
\end{tabular}
\caption{\textbf{Results behind Fig.~\ref{fig:solver_sota}: Fit-Path and LiSM-adapted solvers on the test set.}
GM: geometric mean of adherence (Adh.) and of preservation (Pres.) over the four tasks; the Hack and Time columns below it are means over the four tasks.
The other columns give the same results per task: adherence, internal adherence (Int.), Hack, preservation, and time in seconds per image.
Control: generator without guidance toward the scored task.}
\label{tab:supp-solver-full}
\endgroup
\end{table*}

\section{Additional Qualitative Results}
\label{append:quali}
\begin{figure*}
    \centering
    \includegraphics[width=1\linewidth]{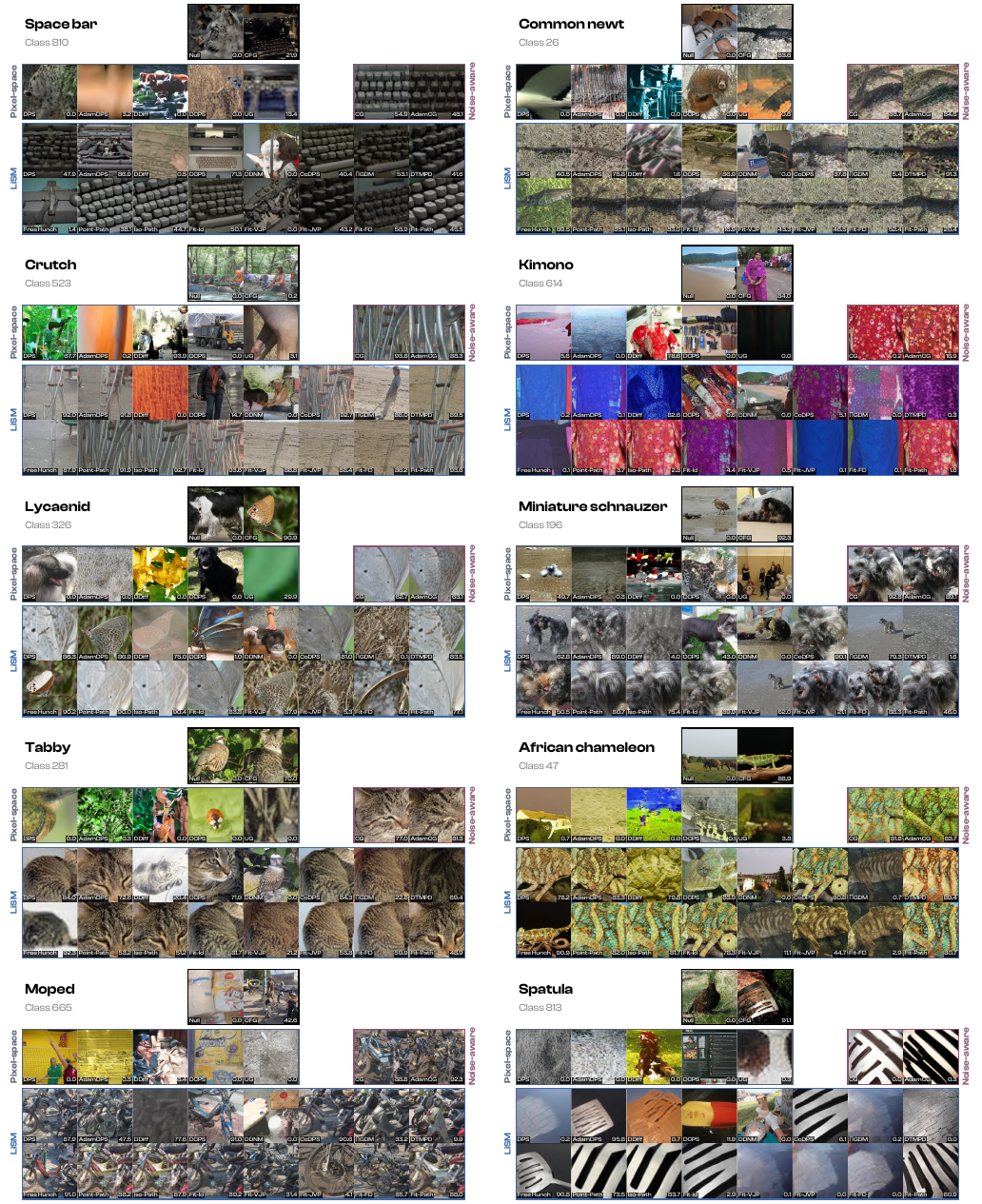}
    \caption{Uncurated qualitative results for Class guidance.}
    \label{fig:supp_quali_class}
\end{figure*}

\begin{figure*}
    \centering
    \includegraphics[width=1\linewidth]{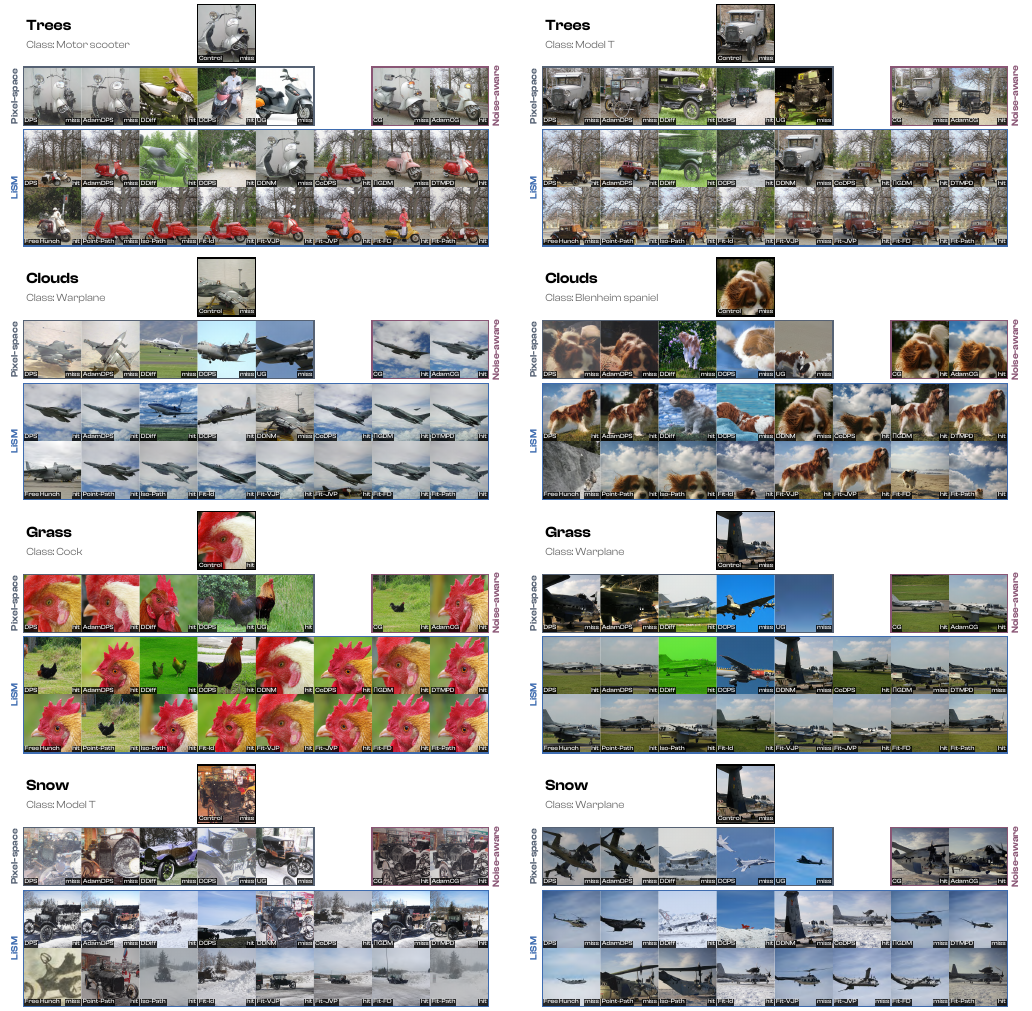}
    \caption{Uncurated qualitative results for Attribute guidance.}
    \label{fig:supp_quali_attributes_1}
\end{figure*}

\begin{figure*}
    \centering
    \includegraphics[width=1\linewidth]{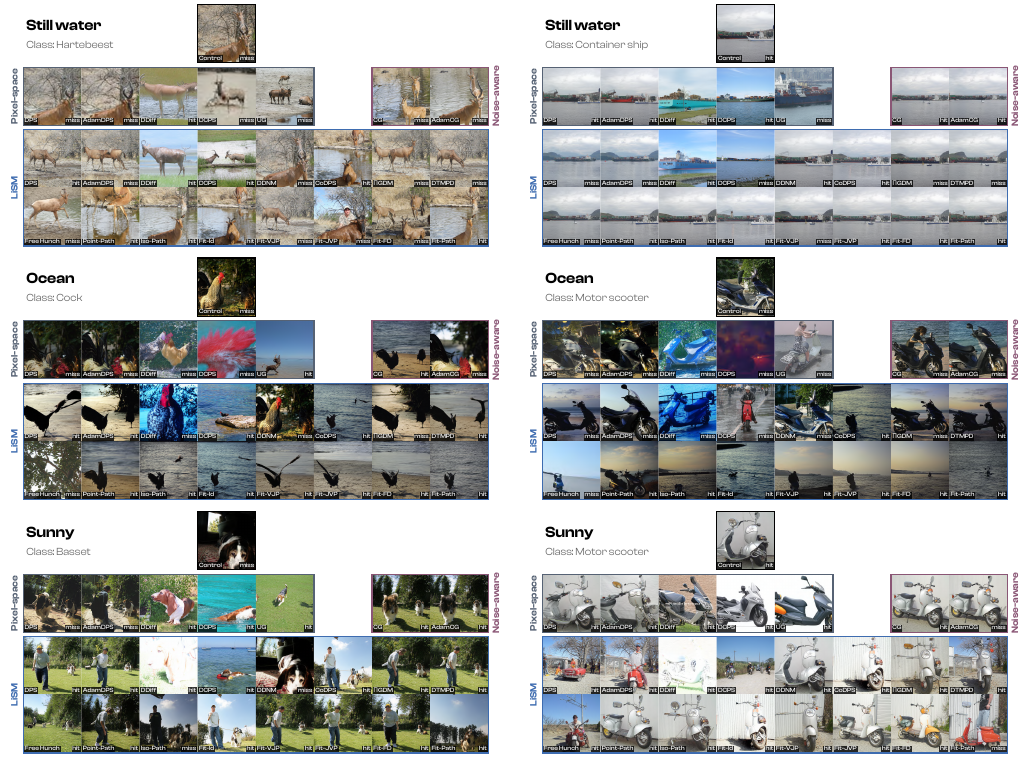}
    \caption{Uncurated qualitative results for Attribute guidance (continued).}
    \label{fig:supp_quali_attributes_2}
\end{figure*}

\begin{figure*}
    \centering
    \includegraphics[width=1\linewidth]{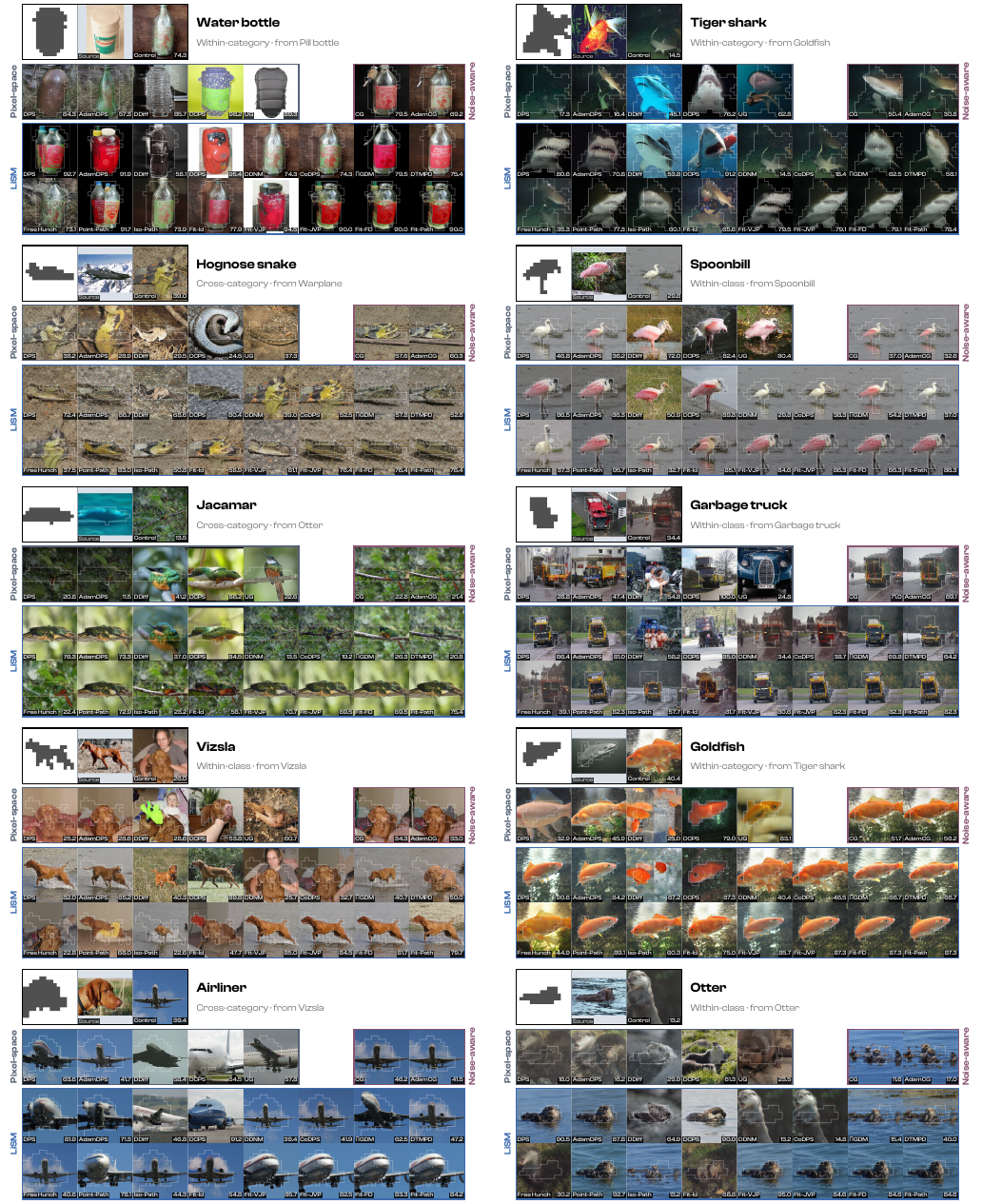}
    \caption{Uncurated qualitative results for Foreground mask guidance.}
    \label{fig:supp_quali_foreground}
\end{figure*}

\begin{figure*}
    \centering
    \includegraphics[width=1\linewidth]{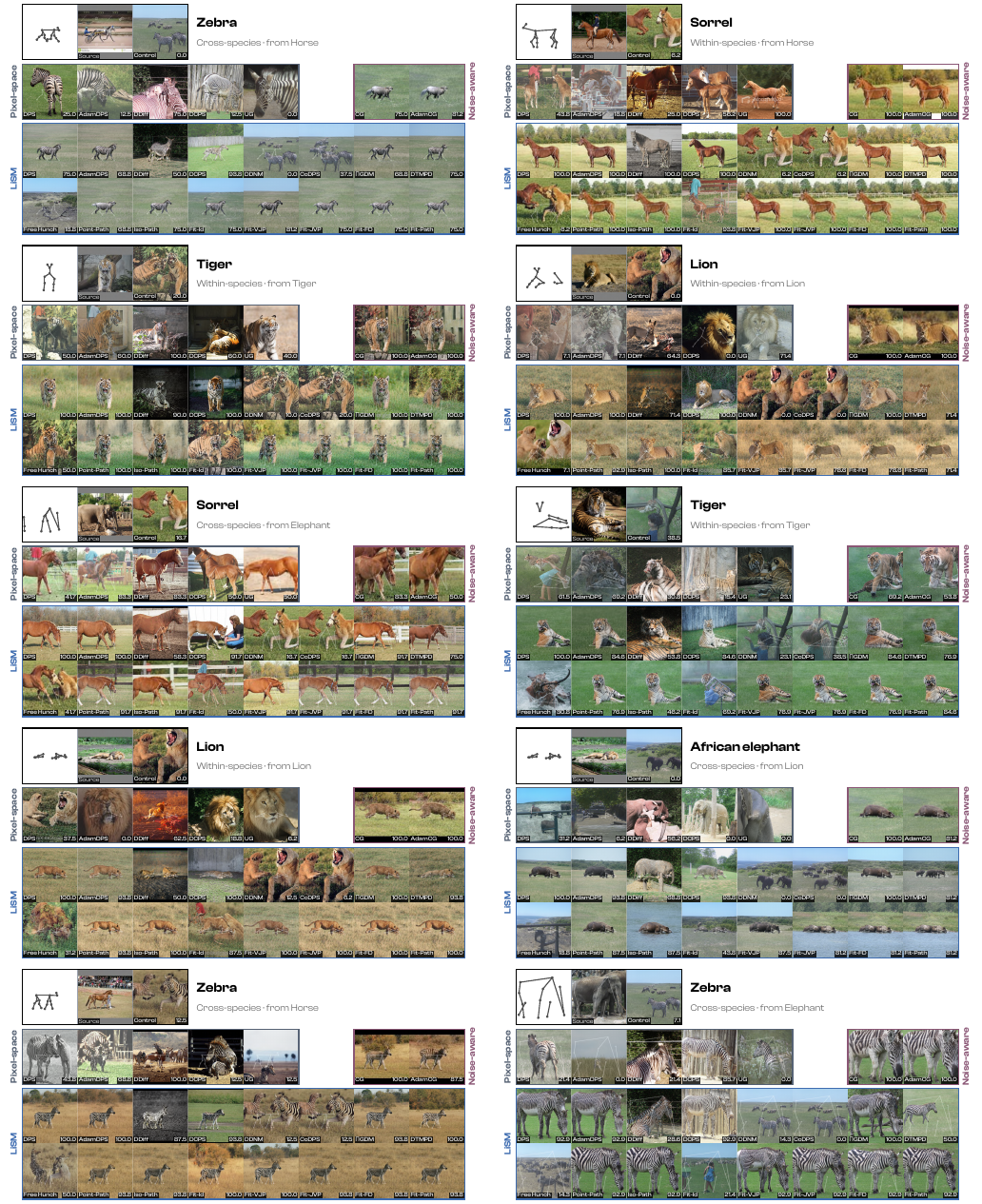}
    \caption{Uncurated qualitative results for Keypoint pose guidance.}
    \label{fig:supp_quali_keypoints}
\end{figure*}

We present uncurated qualitative results for all methods and solvers on class, attributes, foreground, and keypoints. All images are randomly drawn from the shared ablation subset. For attributes, we show two images per attribute, each selected at random. For foreground and keypoints, we first randomly draw the request, then the seed. Results are shown in Fig.~\ref{fig:supp_quali_class} (class), Figs.~\ref{fig:supp_quali_attributes_1} and~\ref{fig:supp_quali_attributes_2} (attributes), Fig.~\ref{fig:supp_quali_foreground} (foreground), and Fig.~\ref{fig:supp_quali_keypoints} (keypoints).


\end{document}